\documentclass[11pt]{article}

\usepackage[preprint]{acl}

\usepackage{times}
\usepackage{latexsym}

\usepackage[T1]{fontenc}

\usepackage[utf8]{inputenc}

\usepackage{microtype}

\usepackage{inconsolata}

\usepackage{algorithm}
\usepackage{algorithmic}
\usepackage{graphicx}
\usepackage{booktabs}
\usepackage{multirow}
\usepackage{tabularx}
\usepackage{multicol}
\usepackage{makecell}
\usepackage{amsmath}
\usepackage{pifont}
\usepackage{subfigure}
\usepackage[most]{tcolorbox}
\usepackage{subcaption}
\usepackage{newfloat}
\usepackage[table,x11names,rgb]{xcolor}

\definecolor{TraditionalBenchmark}{RGB}{0,130,180}
\definecolor{ImageUnderstanding}{RGB}{173,216,230}
\definecolor{DiagnosticBenchmark}{RGB}{255,215,0}
\definecolor{ComprehensiveDiagnosis}{RGB}{255,255,255}
\definecolor{ImageUnderstand}{RGB}{224, 242, 247}
\definecolor{DiseaseKnowledge}{RGB}{240, 255, 240}
\newcommand{\cmark}{\ding{51}}
\newcommand{\xmark}{\ding{55}}
\title{MedReaMM: Evaluating Large Multimodal Models on Expert-Level Clinical Diagnostic Synthesis}

\author{
  \textbf{Lai Wei\textsuperscript{1}\thanks{Equal contribution.}},
  \textbf{Yuchao Chen\textsuperscript{1}\footnotemark[1]},
  \textbf{Zhenbiao Cao\textsuperscript{1}},
  \textbf{Xiaojin Zhang\textsuperscript{1}}, \\
  \textbf{Zhongyu Wei\textsuperscript{2}},
  \textbf{Bangting Wang\textsuperscript{3}},
  \textbf{Wei Chen\textsuperscript{1}\thanks{Corresponding author. \texttt{lemuria\_chen@hust.edu.cn}}},
  \textbf{Xiang Bai\textsuperscript{1}}
  \\[0.5em]
  \textsuperscript{1}Huazhong University of Science and Technology,
  \textsuperscript{2}Fudan University, \\
  \textsuperscript{3}The First Affiliated Hospital with Nanjing Medical University and Jiangsu Province Hospital \\
  \textbf{Code:} \url{https://github.com/thomaswei-cn/MedReaMM}
}

\begin{document}
\maketitle
\begin{abstract}
The application of Large Language Models (LLMs) to diagnostic decision-making has garnered growing interest. However, existing benchmarks largely focus on textual reasoning or isolated visual question-answering (VQA) tasks, lacking holistic integration of clinical narratives and medical imaging, and thus failing to assess the \textit{multimodal diagnostic synthesis} capability central to expert clinical judgment. 
To bridge this gap, we introduce \textbf{MedReaMM}, a benchmark specifically designed to evaluate models' ability to synthesize heterogeneous clinical evidence consisting of detailed patient histories alongside multiple medical images into accurate differential diagnoses under a \textit{complete-information} paradigm.
Constructed from case reports sourced from top-tier medical journals and curated clinical case databases, MedReaMM comprises 625 expert-validated cases with an average of 2.79 medical images per case and a total of 1,042 standardized diagnoses annotated with ICD-11 codes. 
These cases predominantly represent rare, atypical, or multi-system presentations that demand expert-level evidence integration beyond routine pattern recognition. 
We evaluate 23 Large Multimodal Models (LMMs) and find that most achieve diagnostic accuracy scores below 50\%, underscoring a substantial gap in multimodal diagnostic synthesis capability. 
Further analysis reveals that medical knowledge proficiency, medical image understanding, and evidence integration are all highly correlated with diagnostic performance.
\end{abstract}

\section{Introduction}

Recent advances in large language models (LLMs) \cite{jaech2024openai,guo2025deepseek,chen2024huatuogpt} and large multimodal models (LMMs) \cite{tu2024towards,saab2024capabilities} have spurred growing interest in their application to medical decision support, particularly in the challenging task of clinical diagnostic inference. 
In practice, physicians integrate patient history, laboratory results, and medical images to prioritize differential diagnoses and arrive at clinical decisions. 
Notably, medical imaging plays an indispensable role as an objective, standardized, and high-information-density source of clinical evidence \cite{armstrong2010diagnostic}. 
For a model to serve as a competent clinical diagnostic support system, it must not only recall medical knowledge but also extract salient cues from heterogeneous textual and visual inputs and synthesize them coherently into accurate diagnostic conclusions — a capacity we term \textit{multimodal diagnostic synthesis}. 
A benchmark that rigorously evaluates this capability is fundamental for guiding methodological advances toward expert-level clinical intelligence.

\begin{figure*}[!h]
    \centering
    \includegraphics[width=0.8\linewidth]{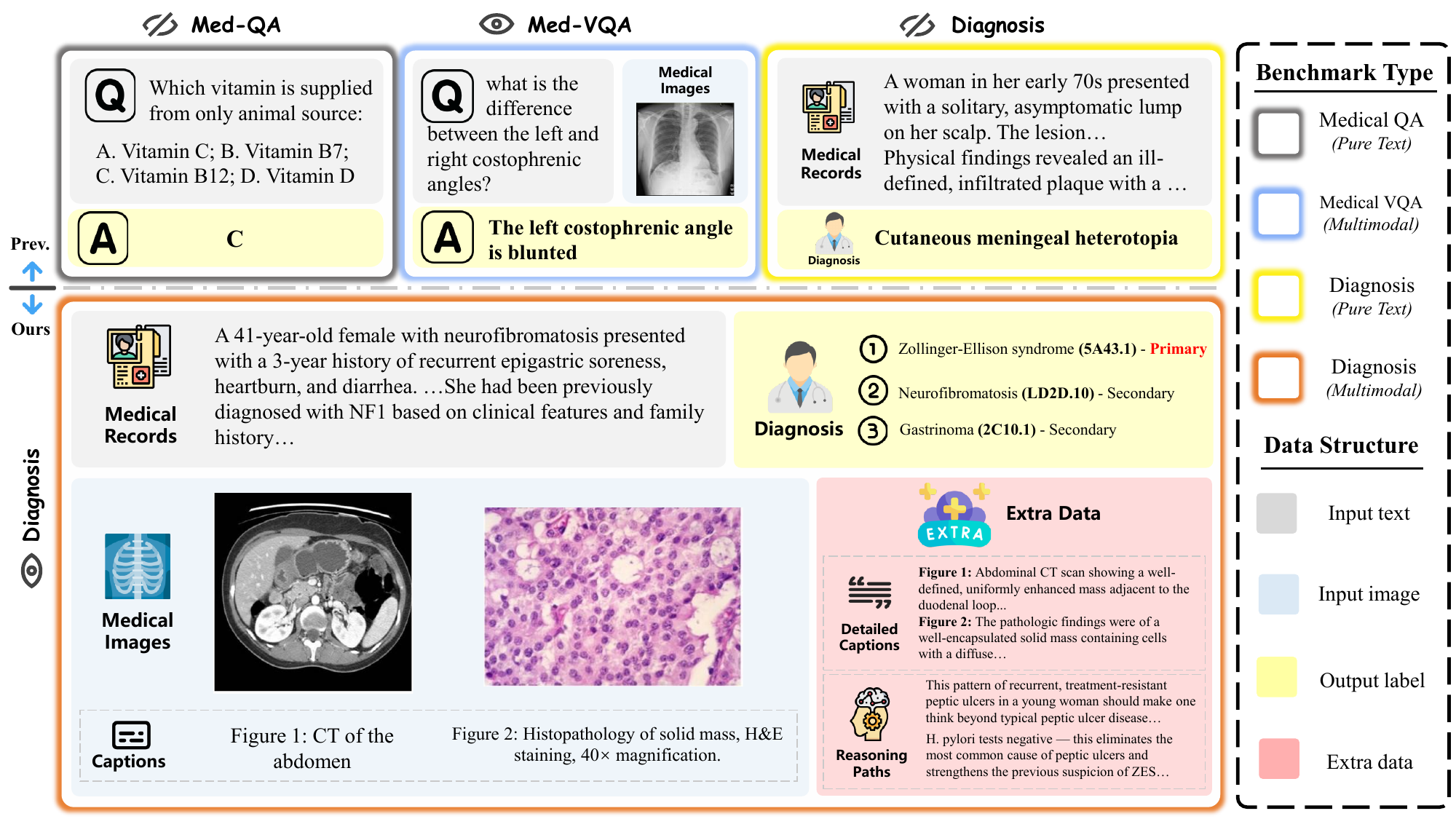}
    \caption{
Comparison between \emph{MedReaMM} and prior benchmarks. 
Upper: Existing datasets often feature factoid-based QA tasks or are limited to text-only input. 
Lower: \emph{MedReaMM} is a diagnostic benchmark featuring multimodal inputs that combine clinical narratives and medical images under a complete-information paradigm, with ICD-11-coded diagnoses enabling standardized evaluation. Beyond core evaluation data, it also includes supplementary data to support in-depth analysis of diagnostic synthesis performance factors.
}
    \label{fig:compare}
\end{figure*}

However, existing benchmarks fall short of capturing these core requirements.
Widely used medical QA and VQA benchmarks \cite{jin2021disease,lau2018dataset} emphasize factual recall or localized perception, thereby failing to capture how heterogeneous evidence is integrated when prioritizing differential diagnoses in high-stakes clinical settings. 
Meanwhile, some diagnostic benchmarks remain restricted to the textual modality \cite{chen2024rarebench,zhu2025diagnosisarena}, overlooking medical images as a critical source of diagnostic evidence. 
A distinct but complementary line of work pursues \textit{interactive} evaluation frameworks in which models iteratively elicit information from patient simulators; while valuable for assessing information-seeking behavior, such frameworks address a fundamentally different capability from diagnostic synthesis under fully assembled evidence. 
This \textit{complete-information} diagnostic synthesis capability, which is the ability to correctly integrate all available clinical data into an accurate diagnosis and a necessary prerequisite before interactive clinical reasoning can be meaningfully built upon, remains inadequately benchmarked.
On the evaluation side, widely adopted multiple-choice formats artificially reduce open-world diagnosis to selection within a predefined candidate space, thereby underestimating the true difficulty of clinical diagnostic inference. 
Conversely, free-form evaluation based solely on string matching \cite{li2023llava} or single-code alignment can misjudge semantically equivalent diagnoses, resulting in biased assessments. 
Together, these limitations highlight the absence of a stable, reproducible, and clinically grounded framework for evaluating diagnostic synthesis proficiency at the frontier of model capability.

To address this gap, we introduce \emph{MedReaMM}, a benchmark designed to challenge LMMs with complete-information, expert-level diagnostic synthesis tasks (Figure~\ref{fig:compare}). 
The complexity of our benchmark are manifested across three dimensions:

\noindent\textbf{Diagnostically Demanding Cases:}\quad Unlike benchmarks composed of standardized medical examination questions, \emph{MedReaMM} comprises 625 expert-validated cases sourced from high-impact clinical journals. Case reports published in such venues characteristically document rare, atypical, or multi-system presentations selected for dissemination precisely because they pose significant diagnostic challenges and carry educational value beyond routine clinical patterns. Such cases demand sophisticated, evidence-driven clinical synthesis that extends well beyond superficial feature matching, and are annotated with 1,042 standardized diagnoses mapped to ICD-11 codes.

\noindent\textbf{Multimodal Synthesis:}\quad To reflect the cognitive demands of integrating fully assembled clinical evidence, models must jointly process detailed clinical narratives alongside diverse medical imaging. To facilitate mechanism analysis, a subset of cases includes reference reasoning paths and detailed image captions. We also apply systematic label-leakage screening with expert review to ensure evaluation remains focused on evidence-based synthesis rather than implicit surface cues.

\begin{table*}[!ht]
\centering
\resizebox{0.8\textwidth}{!}{%
\renewcommand{\arraystretch}{0.95}
\setlength{\tabcolsep}{6pt}
\begin{tabular}{lcccccccccc}
\toprule
\multirow{2}{*}[-0.5em]{\textbf{Benchmark}} & \multirow{2}{*}[-0.5em]{\textbf{Size}} & \multicolumn{2}{c}{\textbf{Image Coverage}} & \multicolumn{2}{c}{\textbf{Text Coverage}} & \multirow{2}{*}[-0.5em]{\makecell[c]{\textbf{Task}\\\textbf{Format}}} & \multirow{2}{*}[-0.5em]{\makecell[c]{\textbf{Data}\\\textbf{Source}}} & \multirow{2}{*}[-0.5em]{\makecell[c]{\textbf{Clinical}\\\textbf{Scenarios}}} & \multirow{2}{*}[-0.5em]{\makecell[c]{\textbf{ICD}\\\textbf{Standardized}}} \\
\cmidrule(r){3-4} \cmidrule(l){5-6}
& & \makecell{\textbf{Average}\\\textbf{Images}} & \makecell{\textbf{Image}\\\textbf{Types}} & \makecell{\textbf{Average}\\\textbf{Length}} & \makecell{\textbf{Clinical}\\\textbf{Info}} & & & & \\
\midrule
\rowcolor{TraditionalBenchmark!15}
MedQA-USMLE \shortcite{jin2021disease} & 1,273 & - & - & 215.46 & Partial & MCQ & Exams & \xmark & \xmark \\ 
\rowcolor{TraditionalBenchmark!15}
MedMCQA-Dev \shortcite{pal2022medmcqa} & 4,183 & - & - & 53.84 & None & MCQ & Exams & \xmark & \xmark \\ 
\rowcolor{TraditionalBenchmark!15}
PubMedQA \shortcite{jin2019pubmedqa} & 1,000 & - & - & 328.41 & Partial & Closed & PubMed & \xmark & \xmark \\ 
\midrule
\rowcolor{ImageUnderstanding!15}
PATH-VQA \shortcite{he2020pathvqa} & 6,719 & 0.13 & 1 & 15.38 & None & Mixed & Textbooks & \xmark & \xmark \\ 
\rowcolor{ImageUnderstanding!15}
VQA-RAD \shortcite{lau2018dataset} & 451 & 0.45 & 3 & 14.61 & None & Mixed & MedPix & \xmark & \xmark \\ 
\rowcolor{ImageUnderstanding!15}
Slake-En \shortcite{liu2021slake} & 1,061 & 0.09 & 3 & 13.97 & None & Mixed & Public & \xmark & \xmark \\ 
\rowcolor{ImageUnderstanding!15}
PMC-VQA \shortcite{zhang2023pmc} & 33,430 & 0.87 & 7 & 61.84 & None & MCQ & PubMed & \xmark & \xmark \\ 
\rowcolor{ImageUnderstanding!15}
OmniMedVQA \shortcite{hu2024omnimedvqa} & 127,995 & 0.92 & 8 & 42.40 & None & Mixed & Public & \xmark & \xmark \\ 
\rowcolor{ImageUnderstanding!15}
MMMU (H\&M) \shortcite{yue2024mmmu} & 1,752 & 1.14 & 8 & 83.56 & None & MCQ & Exams & \xmark & \xmark \\ 
\rowcolor{ImageUnderstanding!15}
MMMU-Pro (H\&M) \shortcite{yue2024mmmupro} & 346 & 1.25 & 8 & 107.08 & None & MCQ & Exams & \xmark & \xmark \\ 
\rowcolor{ImageUnderstanding!15}
MedXpertQA-MM \shortcite{zuo2025medxpertqa} & 2,000 & 1.43 & 8 & 149.35 & Partial & MCQ & Exams & \cmark & \xmark \\ 
\rowcolor{ImageUnderstanding!15}
GMAI-MMBench \shortcite{ye2024gmai} & 21,281 & 0.99 & 9 & 49.85 & None & MCQ & Public & \xmark & \xmark \\ 
\midrule
\rowcolor{DiagnosticBenchmark!15}
CMB-Clin \shortcite{wang2023cmb} & 74 & - & - & 792.55 & Full & Open & Hospital & \cmark & \xmark \\ 
\rowcolor{DiagnosticBenchmark!15}
DiagnosisArena \shortcite{zhu2025diagnosisarena} & 1,113 & - & - & 545.02 & Full & Mixed & Journals & \cmark & \xmark \\ 
\rowcolor{DiagnosticBenchmark!15}
RareArena \shortcite{RareArena} & 72,661 & - & - & 310.36 & Full & Open & PubMed & \cmark & \xmark \\ 
\midrule
\rowcolor{ComprehensiveDiagnosis!15}
\textbf{MedReaMM (Ours)} & 625 & \textbf{2.79} & \textbf{13} & \textbf{833.37} & Full & Open & Journals & \cmark & \cmark \\ 
\bottomrule
\end{tabular}%
}
\caption{Comparison with existing multimodal medical benchmarks. \emph{MedReaMM} demonstrates high complexity through its exceptional multimodal diagnostic design, featuring the highest average image count (2.79 per case), the largest number of imaging modalities (13 types), and comprehensive clinical information (833 words per case). Benchmarks are categorized by their primary focus: \colorbox{TraditionalBenchmark!15}{traditional text-only QA},
\colorbox{ImageUnderstanding!15}{medical image understanding},
\colorbox{DiagnosticBenchmark!15}{diagnostic reasoning}, and \textbf{comprehensive multimodal clinical diagnosis}. }
\label{tab:compare_benchmarks}
\end{table*}

\noindent\textbf{Unconstrained Diagnostic Generation:}\quad To preserve the inherent difficulty of clinical diagnostic inference, \emph{MedReaMM} replaces simplified multiple-choice paradigms with the generation of ranked differential diagnosis lists. To evaluate these open-ended outputs, we adopt primary and complete diagnosis recall perspectives, utilizing a two-stage matching framework that combines ICD-11 code alignment with semantic similarity to mitigate naming and normalization variability.


We evaluate 23 large multimodal models across open-source and proprietary settings. 
Even under a lenient Top-10 criterion, most models achieve below 50\% accuracy, underscoring a significant gap in complete-information multimodal diagnostic synthesis. 
Further analyses show that diagnostic performance strongly correlates with medical knowledge, image understanding, and multimodal evidence integration, while reasoning prompts offer only limited gains constrained by model scale.

\section{Related Work}

\noindent\textbf{Medical Benchmarks for LLMs and LMMs} \quad
Early medical benchmarks mainly focused on factual recall (e.g., MedQA \cite{jin2021disease}, PathVQA \cite{he2020pathvqa} and RJUA-MedDQA \cite{liu2024rjua}).
Recently, medical evaluation has branched into two complementary directions. \textbf{Dynamic} benchmarks, such as \emph{AI Hospital} \cite{fan-etal-2025-ai}, \emph{Agent Hospital} \cite{li2024agent} and \emph{AgentClinic} \cite{schmidgall2024agentclinic}, focus on the iterative process of information seeking and patient interaction \cite{chen-etal-2026-speechmedassist, Chen2022ABF}. In contrast, \textbf{Static} benchmarks focus on the depth of reasoning under complex inputs. While existing static benchmarks are often limited to textual modalities \cite{chen2024rarebench, zhu2025diagnosisarena} or simplified VQA tasks that remain primarily centered on assessing models’ memorization of medical knowledge \cite{zhang2024gmai, zuo2025medxpertqa,Bao2026MedRCubeAM}, \emph{MedReaMM} fills the gap by providing a multimodal, expert-level evaluation of evidence synthesis. 
We view these two paradigms as complementary: dynamic benchmarks evaluate "how to ask," while \emph{MedReaMM} evaluates "how to conclude" when faced with rich, heterogeneous clinical data.
Detailed comparisons with previous benchmarks are summarized in Table~\ref{tab:compare_benchmarks}.



\begin{figure*}[!h]
    \centering
    \includegraphics[width=0.8\linewidth]{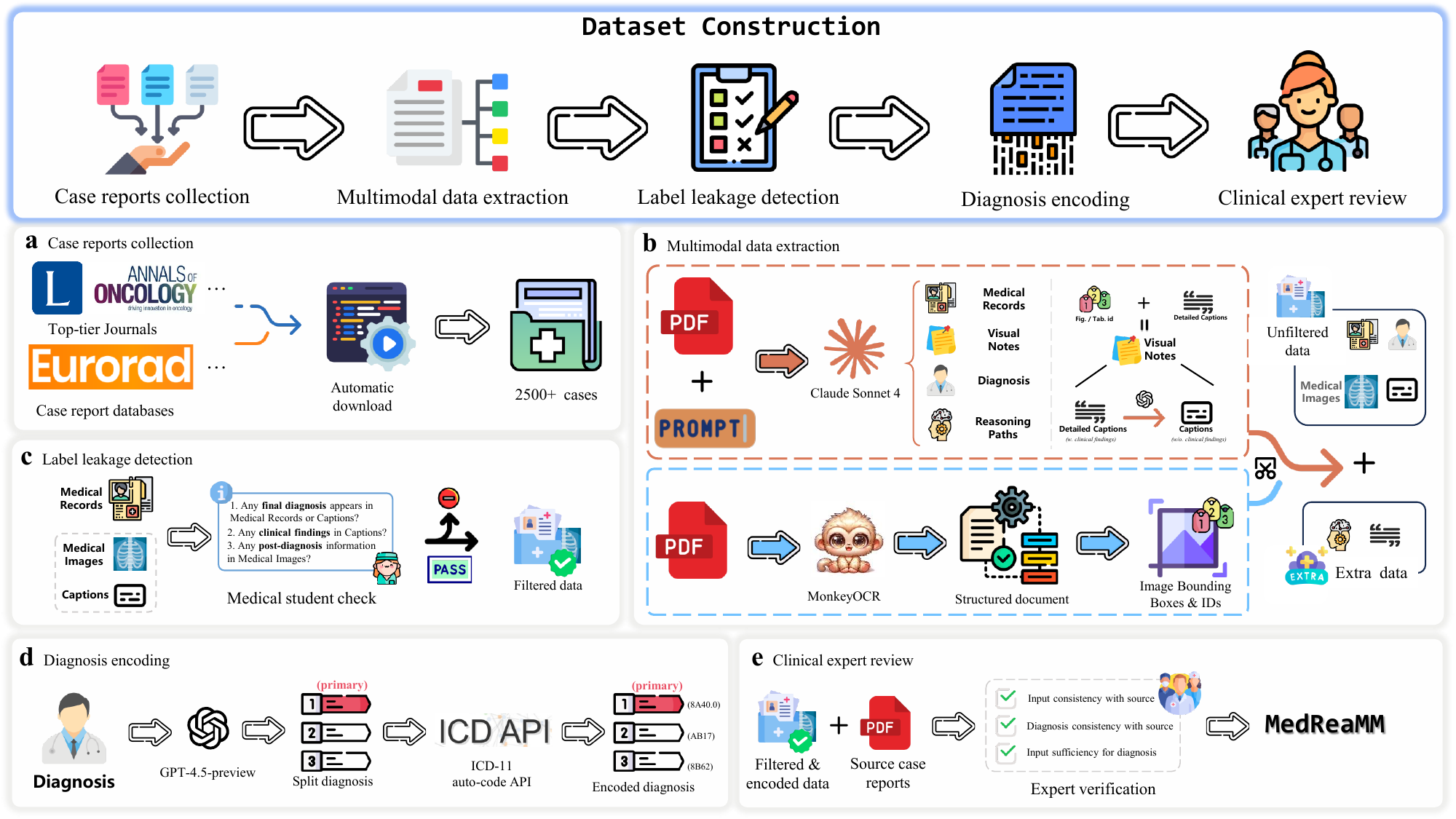}
    \caption{
Overview of the \emph{MedReaMM} construction pipeline. 
\textbf{(a)} \textbf{Case Collection}: Diagnostic case reports are collected from NEJM, The Lancet, Eurorad, and other clinical sources. 
\textbf{(b)} \textbf{Multimodal Extraction}: Clinical text, medical images, and a diverse range of additional information are parsed from case reports.
\textbf{(c)} \textbf{Label Leakage Detection}: A screening process identifies potential label leakage in both textual and visual modalities.
\textbf{(d)} \textbf{Diagnosis Encoding}: All diagnoses are first decomposed into individual disease terms, then standardized via the ICD-11 API.
\textbf{(e)} \textbf{Expert Review}: Final samples are validated by licensed clinicians to ensure quality and clinical plausibility.
}
    \label{fig:construction_diagram}
\end{figure*}

\noindent\textbf{LMMs for Clinical Applications} \quad With their strong vision-language capabilities, LMMs have increasingly been adapted for medicine. Early models like LLaVA-Med \cite{li2023llava} and Med-Flamingo \cite{moor2023med} fine-tuned general LMMs on PubMed and textbook data but were limited by data quality. Later models such as Huatuo-GPT-Vision \cite{chen2024huatuogptvision}, Lingshu \cite{xu2025lingshu}, and MedGemma \cite{sellergren2025medgemma} improved performance by expanding dataset scale and diversity.

Recent works also explore agent-based frameworks integrating LMMs into clinical workflows. MC-CoT \cite{wei2024mc} uses LLM knowledge to guide visual feature extraction, MMedAgent \cite{li2024mmedagent} invokes specialized tools via multi-agent setups, and MDAgent \cite{kim2024mdagents} simulates multidisciplinary consultations.

Despite progress, most models are evaluated on simplified tasks lacking diagnostic complexity. 
In contrast, our benchmark focuses on complex cases with open-ended diagnostics, offering a more rigorous and clinically relevant assessment where AI assistance is most needed.

\section{Benchmark Construction}
This section outlines the benchmark construction pipeline, including case report collection, data structuring, label leakage detection, diagnosis encoding, and validation.

\noindent\textbf{Case Reports Collection} \quad We collect diagnostic cases from two primary sources: \textbf{1) top-tier medical journals} (e.g., NEJM, The Lancet) with an impact factor above 40, and \textbf{2) case-report websites} manually curated for content completeness. 
The full pipeline is illustrated in Figure \ref{fig:construction_diagram}a.

Each case report typically consists of several structured sections, including: \emph{patient history}, \emph{presenting symptoms}, \emph{diagnostic examinations} (e.g., imaging, labs), \emph{clinical reasoning process}, \emph{final diagnosis}, and \emph{treatment or outcome}. 
Many reports also include one or more medical images with descriptive captions that support the diagnostic process.
Given the peer-review mechanisms of journals and case-report platforms, we consider the collected data to be clinically validated, containing diagnostically sufficient information and often involving rare or diagnostically challenging conditions (see Appendix \ref{appendix:data_scope} for discussion on dataset scope).
We collect 2,359 cases from medical journals and 508 from web-based platforms.

\noindent\textbf{Multimodal Data Extraction} \quad To extract diagnostic information and relevant images from PDFs, we use a collaborative pipeline combining Claude Sonnet 4 and MonkeyOCR \cite{li2025monkeyocr}, a state-of-the-art PDF parsing model that converts documents into structured text and images, as illustrated in Figure~\ref{fig:construction_diagram}b.
Specifically, for each case report, we apply Claude Sonnet 4 to extract the following structured elements:
\textbf{1) Medical Records} \( R \), the patient's history, symptoms, and test results that is presented in text format in the original case report;
\textbf{2) Visual Notes} \( N = \{(i_j, C_{d_j})\}_{j=1}^n \), where \( i_j \) denotes the image ID and \( C_{d_j} \) is the original detailed caption;
\textbf{3) Diagnosis} \( D \), the final diagnosis; 
\textbf{4) Reasoning Path} \( P \), optionally extracted for a subset of cases, which traces the logical path from symptoms to diagnosis.

We further process each detailed caption \( C_{d_j} \) using GPT-5 to remove diagnostic interpretations, yielding a purified visual metadata caption \( C_j \). 
These captions retain essential visual metadata, including imaging modality, anatomical region, staining method and magnification level. 
The resulting metadata set is denoted \( C = \{C_j\}_{j=1}^n \).

On the other hand, we use MonkeyOCR to analyze the PDF layout and extract bounding box coordinates \( B_j \) for each image ID \( i_j \):
\[
(i_j, B_j) \in \text{MonkeyOCR}(\text{PDF})
\]

We crop the visual region from the PDF using \( B_j \), yielding the actual medical image \( I_j \). Each image is paired with its filtered caption \( C_j \), forming \( (I_j, C_j) \). 
For model evaluation, we construct a standardized benchmark input:
\[
\mathcal{M}_{\text{eval}} = \left( R,\; \{(I_j, C_j)\}_{j=1}^n,\; D \right)
\]

The original detailed captions \( C_d = \{C_{d_j}\}_{j=1}^n \) and the reasoning path \( P \) are retained as \emph{extra data} for auxiliary analysis and potential future tasks.

\noindent\textbf{Label Leakage Detection} \quad To reduce potential label leakage, we conduct careful screening for each case, as shown in Figure~\ref{fig:construction_diagram}c. 
Three medical-major students review the cases using the original PDFs as reference, checking for leakage from three perspectives:
1) \textbf{Diagnosis mentioned:} whether the primary diagnosis appears in the medical records or image captions;
2) \textbf{Interpretive captions:} whether any caption includes clinical findings or subjective interpretations;
3) \textbf{Post-diagnosis visuals:} whether the image itself contains post-diagnosis information (e.g., recovery status).

Cases flagged by two or more annotators are removed entirely; non-essential images are excluded from otherwise valid cases.
Inter-annotator agreement was assessed using Cohen’s Kappa, yielding a value of 0.6809, indicating substantial agreement. 
Specifically, in our setting, three annotators independently provided binary judgments for each case (leakage vs.\ non-leakage). 
We compute pairwise Cohen’s Kappa scores across annotator pairs and report their average, following the standard formulation $\kappa = \frac{p_o - p_e}{1 - p_e}$, where $p_o$ denotes the observed agreement and $p_e$ the agreement expected by chance \cite{cohen1960coefficient}.
Details of the annotation process are provided in Appendix \ref{appendix:human_anno}.

\noindent\textbf{Diagnosis Encoding} \quad To enable standardized evaluation, we first use GPT-5 to decompose each diagnosis into distinct disease entities. 
This step is necessary because many cases involve multiple coexisting conditions. 
One entity is assigned as the primary diagnosis and the others as secondary. 
Each item is then mapped to ICD-11 codes using the WHO ICD-API\footnote{\url{https://icd.who.int/icdapi}}. 
Unmatched cases are revised and reprocessed with expert input.
Details of the diagnosis normalization and coding process are provided in Appendix~\ref{appendix:icd}.

\noindent\textbf{Clinical Expert Review} ~ {To ensure clinical validity, two senior physicians conducted a final review of all cases (Figure~\ref{fig:construction_diagram}e). 
Each case was evaluated along three axes: \textbf{1)} whether the input matched the original source, \textbf{2)} whether the label diagnosis was faithful, and \textbf{3)} whether the input provided sufficient information to support the diagnosis.
Cases flagged as insufficient or inaccurate by either expert were removed. 
In total, 625 high-quality cases were retained, comprising 70.7\% from peer-reviewed medical journals and 29.3\% from publicly available clinical-case databases and teaching archives, with inter-expert agreement measured by Cohen’s Kappa reaching 0.7917.
Evaluation details are provided in Appendix \ref{appendix:human_anno}

\begin{table*}
\centering
\small
\setlength{\tabcolsep}{2.5pt}
\renewcommand{\arraystretch}{0.9}
\resizebox{0.85\textwidth}{!}{%
\begin{tabular}{l cc cc cc cc cc cc cc}
\toprule
\multirow{3}{*}[-1.5ex]{\textbf{Models}} & \multicolumn{4}{c}{\textbf{Top-5}} & \multicolumn{4}{c}{\textbf{Top-10}} & \multicolumn{2}{c}{\multirow{2}{*}[-1.5ex]{\textbf{Top-1}}} & \multicolumn{2}{c}{\multirow{2}{*}[-1.5ex]{\textbf{MRR}}} & \multicolumn{2}{c}{\multirow{2}{*}[-1.5ex]{\textbf{NDCG@10}}} \\
\cmidrule(lr){2-5} \cmidrule(lr){6-9}
& \multicolumn{2}{c}{\textbf{Primary}} & \multicolumn{2}{c}{\textbf{Complete}} & \multicolumn{2}{c}{\textbf{Primary}} & \multicolumn{2}{c}{\textbf{Complete}} & \multicolumn{2}{c}{} & \multicolumn{2}{c}{} & \multicolumn{2}{c}{} \\
\cmidrule(lr){2-3} \cmidrule(lr){4-5} \cmidrule(lr){6-7} \cmidrule(lr){8-9} \cmidrule(lr){10-11} \cmidrule(lr){12-13} \cmidrule(lr){14-15}
& \textbf{ICD} & \textbf{Sem.} & \textbf{ICD} & \textbf{Sem.} & \textbf{ICD} & \textbf{Sem.} & \textbf{ICD} & \textbf{Sem.} & \textbf{ICD} & \textbf{Sem.} & \textbf{ICD} & \textbf{Sem.} & \textbf{ICD} & \textbf{Sem.} \\
\midrule
Qwen2.5-VL-7B & 22.4 & 17.9 & 13.3 & 9.1 & 23.2 & 18.4 & 14.1 & 9.3 & 17.1 & 13.1 & 19.7 & 15.3 & 20.9 & 16.5 \\
Lingshu-7B & 27.5 & 21.4 & 18.2 & 12.3 & 30.6 & 23.4 & 21.4 & 13.4 & 18.6 & 8.3 & 22.6 & 18.2 & 24.9 & 19.7 \\
MedGemma-4B & 28.3 & 20.8 & 17.6 & 11.2 & 31.2 & 22.9 & 19.7 & 12.6 & 15.8 & 13.3 & 19.6 & 15.9 & 21.3 & 18.3 \\
Qwen3-VL-8B-Instruct & 35.2 & 27.0 & 21.9 & 15.8 & 39.2 & 29.4 & 24.8 & 17.0 & 25.4 & 20.2 & 30.2 & 23.4 & 32.0 & 25.7 \\
GPT-4.1-nano & 37.6 & 26.9 & 24.5 & 17.9 & 41.0 & 28.3 & 28.2 & 19.4 & 25.9 & 18.6 & 30.6 & 22.0 & 32.8 & 23.5 \\
GPT-4o-mini & 36.5 & 28.4 & 23.4 & 16.9 & 39.4 & 30.3 & 27.1 & 18.8 & 23.0 & 17.9 & 28.4 & 21.9 & 31.1 & 24.3 \\
Qwen2.5-VL-32B & 37.6 & 29.0 & 25.0 & 18.4 & 40.6 & 29.9 & 27.2 & 19.2 & 27.2 & 21.8 & 31.6 & 24.8 & 33.7 & 25.9 \\
Lingshu-32B & 36.8 & 30.4 & 23.2 & 16.8 & 39.7 & 33.1 & 25.9 & 18.7 & 27.7 & 23.2 & 31.8 & 26.6 & 33.5 & 28.0 \\
MedGemma-27B & 40.8 & 32.0 & 26.2 & 18.9 & 43.7 & 33.8 & 28.0 & 20.6 & 29.3 & 23.0 & 34.2 & 26.7 & 36.0 & 28.5 \\
LLaMA-4-Scout & 41.8 & 34.2 & 27.8 & 21.6 & 48.3 & 39.0 & 32.8 & 25.0 & 29.3 & 23.7 & 34.9 & 28.4 & 37.7 & 31.7 \\
LLaMA-4-Maverick & 42.7 & 34.2 & 29.1 & 22.4 & 47.8 & 39.7 & 33.3 & 25.8 & 29.4 & 23.5 & 35.1 & 28.3 & 37.9 & 31.8 \\
Qwen2.5-VL-72B & 41.4 & 36.3 & 27.8 & 22.1 & 43.5 & 38.6 & 29.0 & 23.5 & 31.0 & 28.5 & 35.4 & 32.0 & 37.4 & 33.4 \\
GPT-4.1-mini & \underline{52.6} & 38.1 & \underline{35.7} & 26.4 & \underline{56.5} & 39.8 & \underline{38.4} & 27.5 & 41.6 & 30.4 & \underline{46.3} & 33.6 & 48.0 & 35.3 \\
GPT-4o & 52.5 & \textbf{45.3} & 35.4 & \textbf{32.6} & 55.7 & \textbf{48.2} & 37.8 & \underline{31.5} & \underline{41.3} & \underline{35.4} & \underline{46.3} & \textbf{39.8} & \underline{48.1} & \textbf{41.9} \\
GPT-4.1 & \textbf{54.2} & \underline{44.3} & \textbf{36.8} & \underline{30.9} & \textbf{58.1} & \underline{46.4} & \textbf{39.4} & \textbf{32.5} & \textbf{42.7} & \textbf{35.5} & \textbf{48.2} & \underline{39.4} & \textbf{49.7} & \underline{41.3} \\
\midrule
Grok-4 & 29.9 & 24.2 & 19.5 & 14.6 & 29.9 & 24.2 & 19.5 & 14.6 & 24.0 & 19.7 & 27.0 & 21.9 & 28.0 & 22.5 \\
GLM-4.1V-9B-Thinking & 39.8 & 33.9 & 25.8 & 20.6 & 44.0 & 35.8 & 28.5 & 22.4 & 28.5 & 25.4 & 33.5 & 29.0 & 35.9 & 31.6 \\
Qwen3-VL-30B-Thinking & 46.4 & 39.2 & 32.3 & 26.4 & 49.6 & 40.8 & 34.6 & 28.0 & 33.9 & 29.6 & 39.2 & 33.7 & 41.0 & 36.2 \\
o3-mini & 55.0 & 41.8 & 37.8 & 28.2 & 59.0 & 44.3 & 39.8 & 29.3 & 42.1 & 31.7 & 47.5 & \textbf{48.9} & 35.9 & 38.1 \\
o1 & 58.1 & 46.4 & 39.0 & 30.9 & 61.3 & 48.0 & 41.6 & 32.2 & 44.0 & 36.0 & 50.0 & 40.3 & 51.7 & 42.7 \\
Claude-Sonnet-4 & 56.0 & 46.7 & 38.1 & 30.7 & 61.9 & 51.0 & 42.2 & 33.8 & 42.4 & 34.1 & 48.4 & 39.3 & 50.8 & 43.0 \\
Gemini-2.5-Flash & \underline{60.0} & \underline{50.1} & \underline{43.0} & \underline{34.1} & \underline{63.2} & \underline{52.2} & \underline{46.1} & \underline{36.3} & \underline{46.6} & \underline{39.5} & \underline{52.4} & 44.2 & \underline{54.3} & \underline{47.0} \\
Gemini-2.5-Pro & \textbf{65.9} & \textbf{54.7} & \textbf{45.8} & \textbf{37.8} & \textbf{68.0} & \textbf{56.5} & \textbf{47.8} & \textbf{39.4} & \textbf{51.5} & \textbf{43.5} & \textbf{57.6} & \underline{48.1} & \textbf{59.5} & \textbf{51.0} \\
\bottomrule
\end{tabular}
}
\caption{Performance of 23 state-of-the-art LMMs on \emph{MedReaMM}. Evaluation metrics include Recall (Top-5/10) and rank-sensitive indices (Top-1, MRR, NDCG@10). Note that Top-1 and MRR exclusively use the Primary diagnosis as the ground truth. All values are scaled by 100 for readability.}
\label{tab:main_exp}
\end{table*}

\section{Experiment}
\subsection{Implementation Details}

To ensure fair comparison, we use identical prompts for all models in both main and supplementary experiments (see Appendix \ref{appendix:prompts}).

In the main experiment, we instruct models to output a ranked list of ten possible diagnoses under the following constraints:  
1) Each item must represent a single, standalone disease. Models are explicitly prohibited from listing multiple diseases in a single entry or describing diagnoses using free-text sentences.  
2) The diagnoses must be sorted in descending order of likelihood, with the most probable disease listed first.
We adopt greedy decoding for all models, in order to minimize output randomness and ensure reproducibility.
Regarding reasoning behavior, we retain each model’s default setting without manual intervention.  
These default behaviors can be broadly categorized into two types:  
1) \textbf{Reasoning-enabled models}, such as Gemini-2.5-Pro\cite{comanici2025gemini} and o1;  
2) \textbf{Standard models}, including GPT-4.1, LLaMA-4, MedGemma-27B \cite{sellergren2025medgemma} and Lingshu-32B \cite{xu2025lingshu}.

\subsection{Evaluation}

We evaluate models under two distinct sets of metrics. The first set focuses on \textbf{Top-$5/10$ Diagnostic Recall}. A case is correctly diagnosed if the ground-truth appears among the model's top-$k$ predictions. Within this set, we define two tasks: \textbf{1) Primary Diagnosis Recall:} success if the primary diagnosis is included in the predictions. \textbf{2) Complete Diagnosis Recall:} success only if all annotated diagnoses (primary and secondary) are covered.

The second set comprises \textbf{Ranking-Sensitive Metrics} to assess diagnostic precision: Top-1 Accuracy, Mean Reciprocal Rank (MRR), and Normalized Discounted Cumulative Gain at 10 (NDCG@10). For Top-1 and MRR, we strictly use the primary diagnosis as the ground truth. MRR is defined as $\frac{1}{|Q|} \sum_{i=1}^{|Q|} \frac{1}{\text{rank}_i}$, where $\text{rank}_i$ is the position of the primary diagnosis. For NDCG@10, we compute the quality of the top-10 list by assigning hierarchical relevance scores: a primary diagnosis match receives a score of 2, while other diagnoses in the complete set receive 1, and mismatches receive 0. The final score is $\text{NDCG}@10 = \frac{\text{DCG}@10}{\text{IDCG}@10}$, where IDCG is calculated based on an ideal ordering of these relevance scores.

To ensure fair evaluation across models with varying linguistic normalization capabilities, we employ two complementary matching approaches. \textbf{1) ICD-11 Code Matching:} Predictions are mapped to ICD-11 codes via the WHO Auto-code API (with Flex-Search fallback), requiring exact code equivalence. \textbf{2) Semantic Similarity Matching:} We compute cosine similarity between SapBERT \cite{liu2020self} embeddings of predictions and ground-truth terms, using an empirical threshold of 0.85 to define a match. Combining these approaches prevents unfair penalties for synonymous medical expressions and focuses the benchmark on diagnostic reasoning.

\begin{figure*}[!h]
    \centering
    \includegraphics[width=0.8\linewidth]{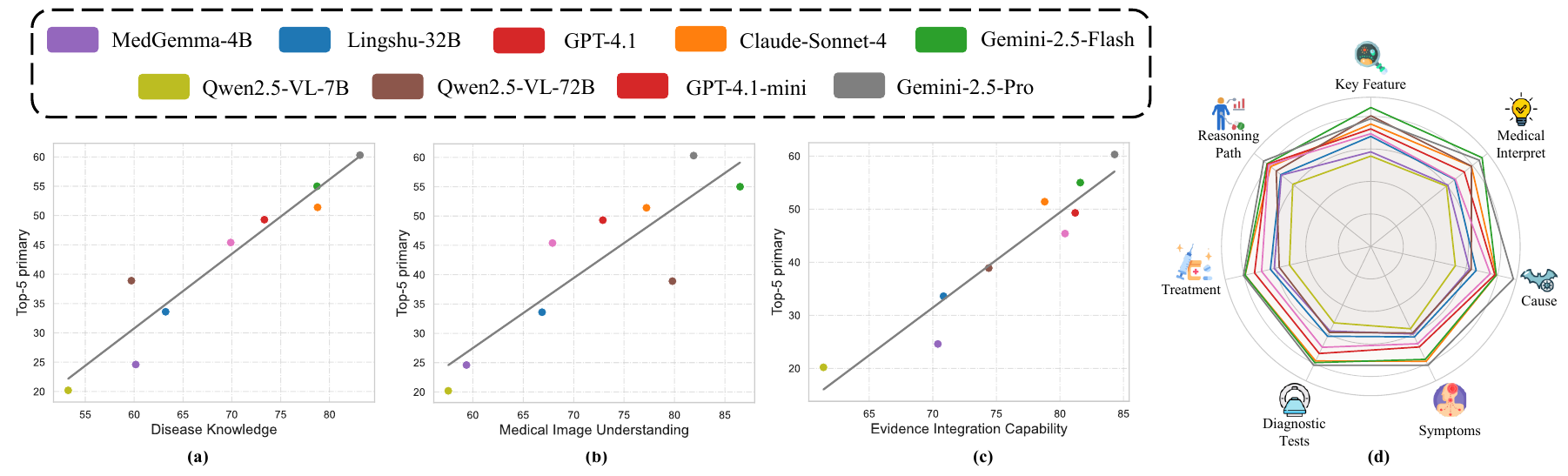}
    \caption{Linear regression analysis between primary diagnosis recall under the top-5 setting and 3 factors: medical image understanding ability, disease knowledge proficiency and evidence integration capability. Correlations are observed in all cases.}
    \label{fig:relation_analysis}
\end{figure*}







\subsection{Main Results}

Table~\ref{tab:main_exp} shows the performance of 23 leading vision-language and text-only models on \emph{MedReaMM}. In our experiment, the best-performing models are Gemini-2.5-Pro and Gemini-2.5-Flash, both exceeding 55\% in the Top-10 setting under the \emph{Primary Diagnosis Recall} task. Notably, even under this most lenient evaluation criterion, the majority of models struggle to surpass the 50\% threshold. This substantial performance gap highlights the current limitations of state-of-the-art models in multimodal clinical diagnostic tasks, underscoring both the difficulty and necessity of our proposed benchmark.

Furthermore, we observe a strong positive correlation between ranking-sensitive metrics and Top-5/10 recall performance.
This correlation suggests that models capable of robust evidence integration not only retrieve relevant differentials but also systematically filter out clinical hallucinations, thereby assigning higher probabilities to the strictly correct diagnoses backed by multimodal evidence.

Additionally, reasoning models consistently outperform non-reasoning models across all metrics. Models equipped with advanced reasoning paradigms achieve top-tier performance on both recall and strict ranking indices, highlighting the critical role of deep clinical reasoning in solving complex diagnostic cases. 
A representative example is provided in Appendix \ref{appendix:case}. To ensure the reliability of these findings, we report confidence intervals and confirm result stability across prompt variations in Appendix \ref{appendix:robust}.

\subsection{Drivers of Diagnostic Performance}

Our main experiments reveal performance gaps across models, motivating an analysis of the factors underlying diagnostic ability. 
Using \emph{MedReaMM} and its auxiliary data, we decompose diagnostic capability into three components: disease knowledge, medical image understanding, and evidence integration capability. 
These dimensions respectively capture what a model knows, what it can perceive, and how it rationalizes heterogeneous clinical evidence. 
For each component, we design an auxiliary evaluation task and evaluate nine representative models to disentangle these driving factors. 
Model outputs are scored by GPT-5, normalized to a 0–100 scale, and analyzed for their impact on Top-5 \emph{Primary Diagnosis Recall}.

\begin{table}[!h]
  \small
  \renewcommand{\arraystretch}{0.95}
  \centering
  \resizebox{0.4\textwidth}{!}{%
  \begin{tabular}{l c c c}
  \toprule
  \textbf{Models} & \textbf{Hint} & \textbf{Top-5 Primary} & \textbf{Top-10 Primary} \\
  \midrule
  Claude-Sonnet-4 &  & 52.2 & 56.4 \\
  \rowcolor{gray!15} Claude-Sonnet-4 & \cmark & 58.9\rlap{$_{\textbf{\scriptsize(+6.7)}}$} &
  61.4\rlap{$_{\textbf{\scriptsize(+5.0)}}$} \\
  Gemini-2.5-Flash &  & 55.8 & 56.4 \\
  \rowcolor{gray!15} Gemini-2.5-Flash & \cmark & 58.6\rlap{$_{\textbf{\scriptsize(+2.8)}}$} &
  60.3\rlap{$_{\textbf{\scriptsize(+3.9)}}$} \\
  GPT-4.1 &  & 53.9 & 55.3 \\
  \rowcolor{gray!15} GPT-4.1 & \cmark & 55.0\rlap{$_{\textbf{\scriptsize(+1.1)}}$} &
  56.7\rlap{$_{\textbf{\scriptsize(+1.4)}}$} \\
  Llama-4-Maverick &  & 53.6& 55.8\\
  \rowcolor{gray!15} Llama-4-Maverick & \cmark & 55.8\rlap{$_{\textbf{\scriptsize(+2.2)}}$} &
  56.4\rlap{$_{\textbf{\scriptsize(+0.6)}}$} \\
  GPT-4o &  & 48.1 & 52.5\\
  \rowcolor{gray!15} GPT-4o & \cmark & 50.6\rlap{$_{\textbf{\scriptsize(+2.5)}}$} &
  52.8\rlap{$_{\textbf{\scriptsize(+0.3)}}$} \\
  Qwen2.5-VL-72B &  & 40.3 & 41.7 \\
  \rowcolor{gray!15} Qwen2.5-VL-72B & \cmark & 45.0\rlap{$_{\textbf{\scriptsize(+4.7)}}$} &
  45.6\rlap{$_{\textbf{\scriptsize(+3.9)}}$} \\
  Qwen2.5-VL-7B &  & 15.3 & 15.3 \\
  \rowcolor{gray!15} Qwen2.5-VL-7B & \cmark & 14.2\rlap{$_{\text{\scriptsize(-1.1)}}$} &
  14.2\rlap{$_{\text{\scriptsize(-1.1)}}$} \\
  \bottomrule
  \end{tabular}
  }
  \caption{Performance comparison on the rare disease subset of 180 cases.}
  \label{tab:rare_disease_performance}
  \end{table}

\noindent\textbf{Disease Knowledge.} \quad We randomly sampled 90 distinct diseases from \emph{MedReaMM}.
A senior physician annotated each disease across four aspects: 1) \emph{Causes}, 2) \emph{Symptoms}, 3) \emph{Diagnostic Tests}, and 4) \emph{Treatments}, based on sources such as the MSD Manual\footnote{\url{https://www.msdmanuals.com/professional}}.
The chosen LMMs were prompted with disease names to generate structured responses along these dimensions. Their outputs were scored against physician annotations.

\noindent\textbf{Medical Image Understanding.} \quad We randomly sampled 100 images, and each image, together with its original caption, was shown to the model, followed by the prompt: \emph{Provide a detailed description of both the visible contents and the information that can be inferred}.
We then assess model outputs using detailed captions from extra data of \emph{MedReaMM} as reference, with evaluation focusing on two aspects: \emph{Key Feature Detection} (accuracy in identifying vital signs) and \emph{Medical Interpretation} (ability to infer clinically relevant information).

\noindent\textbf{Evidence Integration Capability.} \quad We assessed this capability using gold-standard reasoning paths included in the extra data of \emph{MedReaMM}. 
For each case with a reference reasoning path, models were given the medical records, associated images and captions, and the ground-truth final diagnosis. 
They were then prompted to generate a reasoning process connecting the evidence to the diagnosis, which is scored against reference paths. 
This sub-task specifically evaluates the model's capacity for \emph{post-hoc rationalization}. 
By providing the ground-truth diagnosis as a fixed constraint in the input, we isolate and measure the model's ability to reconstruct the logical bridge between clinical evidence and a specific conclusion, thereby assessing its interpretive depth independently of predictive accuracy.

Figure~\ref{fig:relation_analysis}d reports the capability scores for the evaluated models, while Figures~\ref{fig:relation_analysis}a--c illustrate their relationships with overall diagnostic performance. 
While general trends show positive correlations across all factors, we identify a \emph{perceptual-cognitive misalignment} in some LMMs. 
For instance, Qwen-2.5-VL-72B exhibits a sharp vision-knowledge mismatch: despite achieving a high Medical Image Understanding score (79.8, near Gemini-2.5-Pro's 81.9), its relatively low Disease Knowledge (59.7) severely bottlenecks its Diagnostic Performance (38.9). 
This anomaly highlights a \emph{weakest-link} phenomenon in multimodal clinical synthesis: superior visual grounding is rendered ineffective without a robust semantic knowledge base to accurately map perceptual findings to pathophysiological etiologies. 
Consequently, these structural bottlenecks imply that simply scaling general vision-language parameters is insufficient for complex diagnostic reasoning. 
Future optimization strategies must directly address domain knowledge deficits, potentially through medical-specific continual pre-training, RAG for clinical facts, or explicitly aligning visual encoders with medical knowledge graphs. 
Prompts and human verifications are described in Appendices~\ref{appendix:prompts} and \ref{appendix:human_eval}.

\begin{table}[htbp]
  \small
  \centering
  \renewcommand{\arraystretch}{0.95}
  \resizebox{0.40\textwidth}{!}{%
  \begin{tabular}{l c c c}
  \toprule
  \textbf{Models} & \textbf{CoT} & \textbf{Top-5 Primary} & \textbf{Top-10 Primary} \\
  \midrule
  GPT-4o &  & 48.9 & 51.9 \\
  \rowcolor{gray!15} GPT-4o & \cmark & 50.3\rlap{$_{\textbf{\scriptsize(+1.4)}}$} &
  53.5\rlap{$_{\textbf{\scriptsize(+1.6)}}$} \\
  GPT-4.1 &  & 49.3 & 52.2 \\
  \rowcolor{gray!15} GPT-4.1 & \cmark & 51.1\rlap{$_{\textbf{\scriptsize(+1.8)}}$} &
  55.1\rlap{$_{\textbf{\scriptsize(+2.9)}}$} \\
  Llama-4-Maverick &  & 38.5 & 43.8 \\
  \rowcolor{gray!15} Llama-4-Maverick & \cmark & 49.0\rlap{$_{\textbf{\scriptsize(+10.5)}}$} &
  49.5\rlap{$_{\textbf{\scriptsize(+5.8)}}$} \\
  Qwen2.5-VL-72B &  & 38.9 & 41.0 \\
  \rowcolor{gray!15} Qwen2.5-VL-72B & \cmark & 41.0\rlap{$_{\textbf{\scriptsize(+2.1)}}$} & 43.1 \rlap{$_{\textbf{\scriptsize(+2.1)}}$}\\
  Lingshu-32B &  & 33.6 & 36.4 \\
  \rowcolor{gray!15} Lingshu-32B & \cmark & 35.7\rlap{$_{\textbf{\scriptsize(+2.1)}}$} &
  37.5\rlap{$_{\textbf{\scriptsize(+1.1)}}$} \\
  Qwen2.5-VL-7B &  & 20.2 & 20.8 \\
  \rowcolor{gray!15} Qwen2.5-VL-7B & \cmark & 11.8\rlap{$_{\text{\scriptsize(-8.3)}}$} &
  11.8\rlap{$_{\text{\scriptsize(-9.0)}}$} \\
  \bottomrule
  \end{tabular}
  }
  \caption{Impact of Chain-of-Thought prompting on standard models across benchmark cases.}
  \label{tab:cot_performance}
  \end{table}

\subsection{Effects of Prompt-Level Interventions}
In this section, we further examine whether prompt-level interventions can improve model behavior without modifying underlying parameters. 
In particular, we study two commonly used strategies in practice: explicit prior knowledge injection and explicit reasoning prompts.

\noindent\textbf{Prior Knowledge Integration} \quad We investigate whether explicit prior knowledge enhances model performance on rare diseases using 180 expert-validated rare disease cases from \emph{MedReaMM}. We compared baseline prompts against prompts with an additional hint: \emph{This case represents a rare medical condition}.
Table~\ref{tab:rare_disease_performance} reveals heterogeneous responses across models. Claude-Sonnet-4 shows the largest improvement, while Qwen2.5-VL-7B exhibits performance degradation, likely due to its limited 7B parameter scale hindering effective clinical reasoning. 

\noindent\textbf{Explicit Reasoning Prompts} \quad We tested all 625 cases, comparing baseline prompts against CoT-guided prompts instructing models to \emph{first provide step-by-step clinical reasoning, then give your final diagnosis}.Table~\ref{tab:cot_performance} reveals significant improvements across most standard models with explicit reasoning guidance. LLaMA-4-Maverick shows the largest gains, followed by Qwen2.5-VL-72B. However, Qwen2.5-VL-7B exhibits performance degradation, reflecting insufficient fundamental capabilities at the 7B parameter scale for complex diagnostic reasoning.

\begin{figure}[!h]
    \centering
    \subfigure[Top-5 Primary]{\label{fig:image_impact_top5}\includegraphics[width=0.4\textwidth]{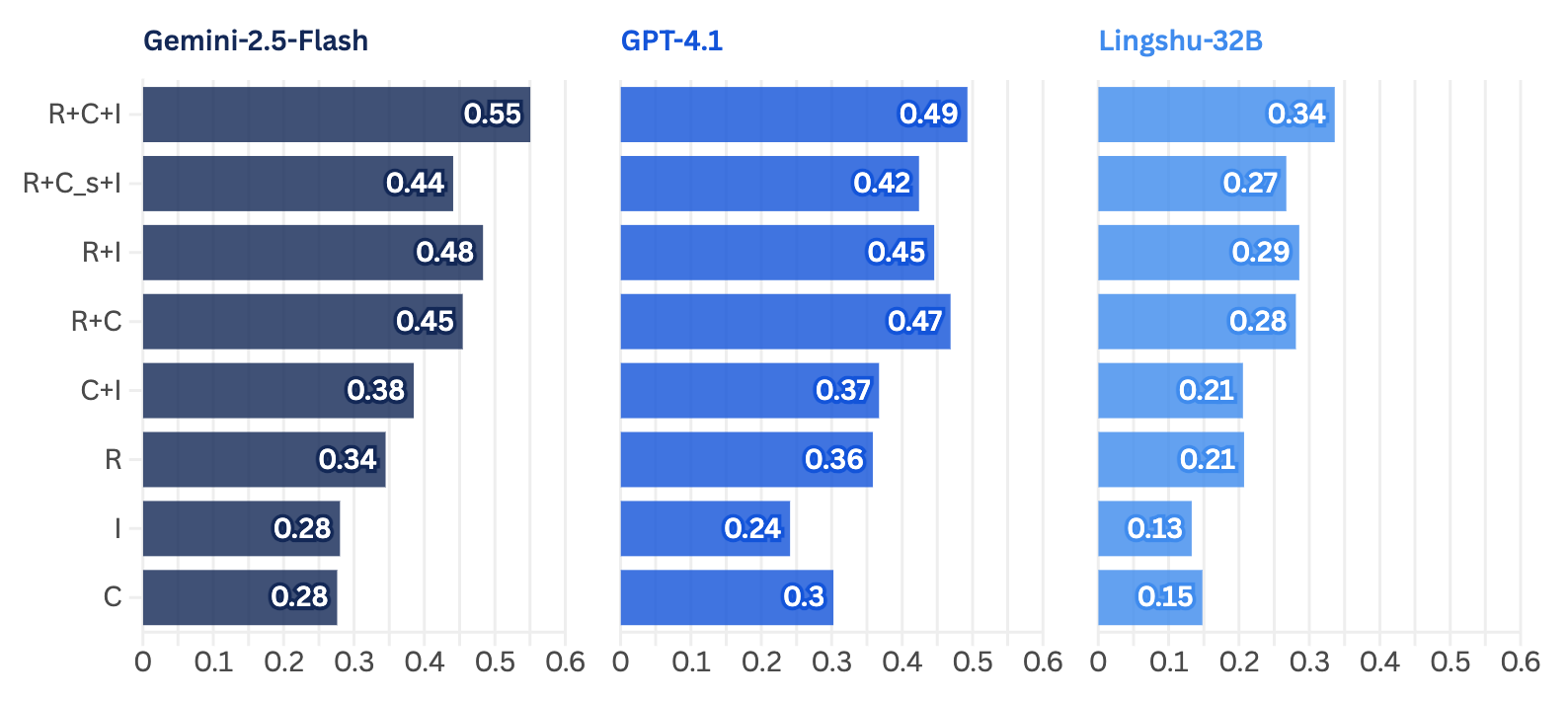}}

    \subfigure[Top-10 Primary]{\label{fig:image_impact_top10}\includegraphics[width=0.40\textwidth]{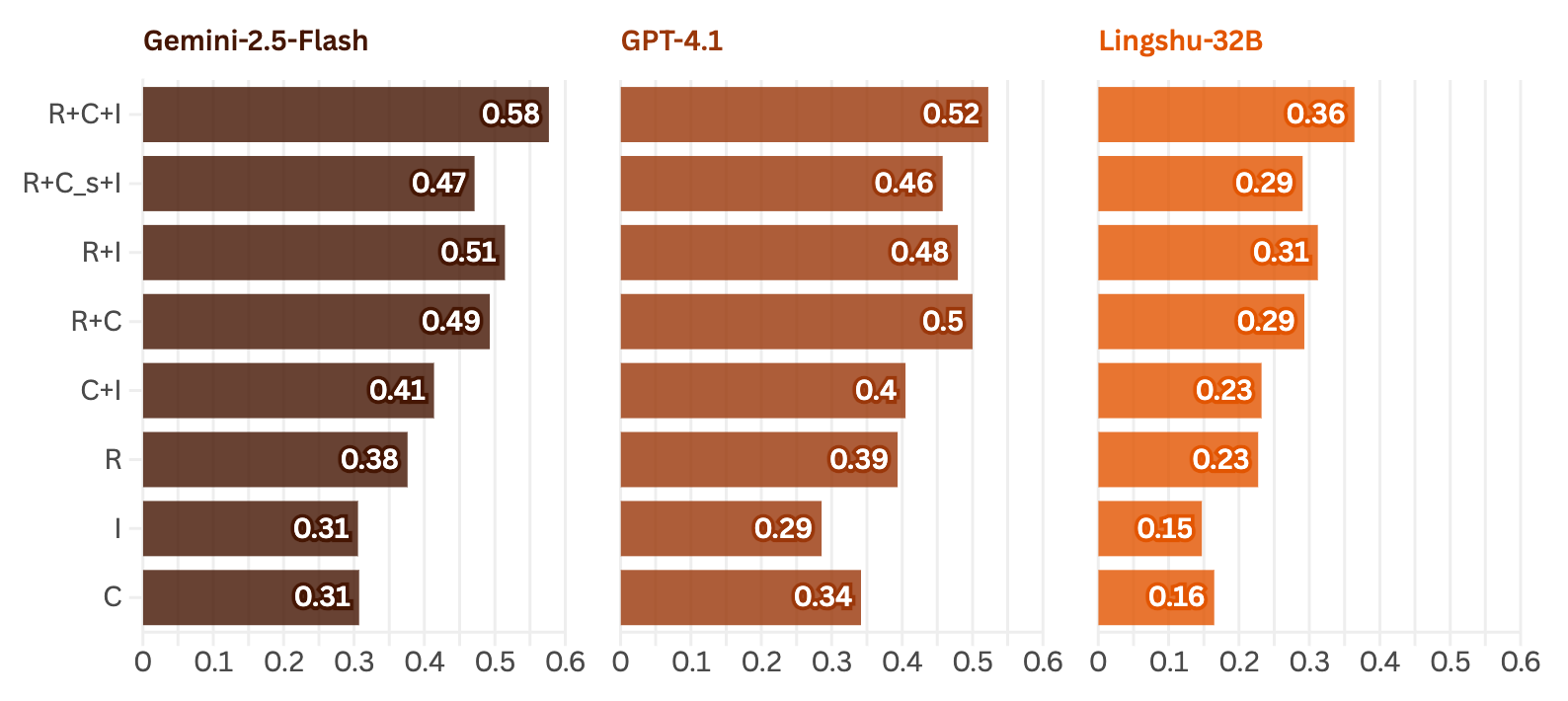}}
    \caption{Ablation results comparing the effect of different input combinations on \emph{Primary Diagnosis Recall}.}
    \label{fig:image_impact}
\end{figure}

These findings demonstrate that these prompt-level interventions demonstrate measurable but uneven effects on diagnostic performance. While explicit priors and reasoning guidance can influence model behavior, their benefits are constrained by underlying capability limitations.

\subsection{Impact of Image and Captions}

To examine how each input component contributes to diagnosis, we conduct an ablation study under seven configurations shown in Figure~\ref{fig:image_impact}: medical records only (\(R\)); images only (\(I\)); records with captions (\(R+C\)); images with captions (\(I+C\)); records with images (\(R+I\)); the full input combining records, images, and aligned captions (\(R+C+I\)); and a control setting where captions are randomly sampled from other cases (\(C_{\text{s}}\)) and paired with the current image, yielding a shuffled-caption configuration (\(R+I+C_{\text{s}}\)).

As shown in Figure~\ref{fig:image_impact}, the full input consistently achieves the best performance across models. Removing any component leads to a performance drop, indicating each modality provides non-redundant evidence and is necessary for reliable multimodal diagnosis.
Notably, the shuffled-caption control substantially underperforms the aligned setting and even the image--record baseline. This suggests that current models rely on captions as a key bridge for interpreting images, highlighting the importance of caption-image alignment.

\section{Conclusion}
We introduce \emph{MedReaMM}, a multimodal benchmark for evaluating diagnostic capabilities of LMMs.
Results reveal that current models remain far from reliable clinical use.
\section*{Limitations}
Despite its strengths, this work has several limitations that should be acknowledged.

First, although \emph{MedReaMM} is constructed from real-world case reports and emphasizes clinically complex diagnostic reasoning, it does not fully capture the interactive and iterative nature of real clinical workflows.
In practice, physicians often refine hypotheses through follow-up questioning, additional examinations, and multidisciplinary consultation.
Our benchmark evaluates diagnosis under a fixed-input setting and therefore cannot assess adaptive decision-making or information-seeking behaviors.

Second, while \emph{MedReaMM} includes diverse imaging modalities and rich textual records, the benchmark relies on curated case reports rather than raw electronic health records.
As a result, the data distribution may differ from that of routine clinical practice, where information is often noisier, incomplete, or inconsistently documented.
This choice improves data quality and diagnostic clarity but may limit direct generalization to real-world hospital systems.

Third, although we take extensive measures to mitigate label leakage, including systematic screening and expert review, it is difficult to entirely rule out subtle forms of implicit cues inherited from case report narratives or image selection.
Such residual signals may influence both human and model performance and remain a broader challenge for diagnostic benchmark construction.

Fourth, our primary evaluation focuses on diagnostic accuracy under open-ended generation, which does not directly measure other clinically relevant aspects such as uncertainty calibration, risk awareness, or downstream clinical impact.
Similarly, while we include a supplementary comparison with human clinicians (detailed in Appendix \ref{appendix:human_vs_lmm}), this evaluation relies on a constrained multiple-choice formulation and should not be interpreted as a comprehensive assessment of human diagnostic competence.

Finally, our analysis centers on current large language and multimodal models.
As model architectures, training data, and reasoning mechanisms continue to evolve, absolute performance levels reported in this work may change.
Nevertheless, we believe that the core challenges highlighted by MedReaMM—namely multimodal evidence integration and clinically grounded reasoning—will remain relevant for future systems.

\bibliography{custom}

\appendix

\begin{center}
    {\LARGE \bfseries Appendix}
\end{center}
\section{Dataset Statistics}

To provide a comprehensive understanding of \emph{MedReaMM}, we present detailed statistics regarding image modalities, disease categories, and diagnostic label distributions. 
These statistics highlight the benchmark's clinical diversity and its alignment with real-world diagnostic scenarios.

\begin{figure*}[!ht]
    \centering
    \subfigure[Image Modality Distribution]{\includegraphics[width=0.45\textwidth]{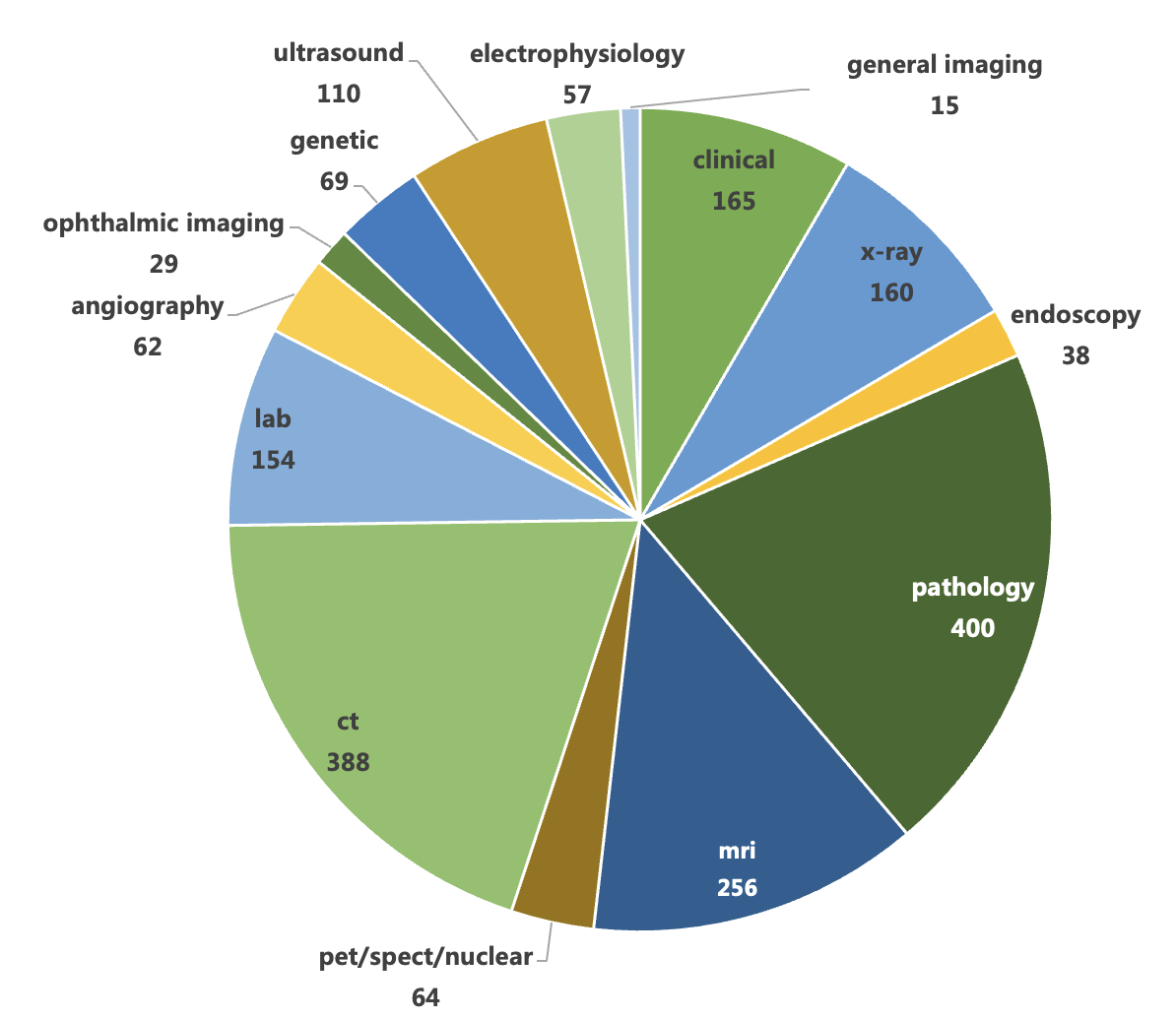}}
    \hfill
    \subfigure[Diagnosis Category Distribution]{\includegraphics[width=0.5\textwidth]{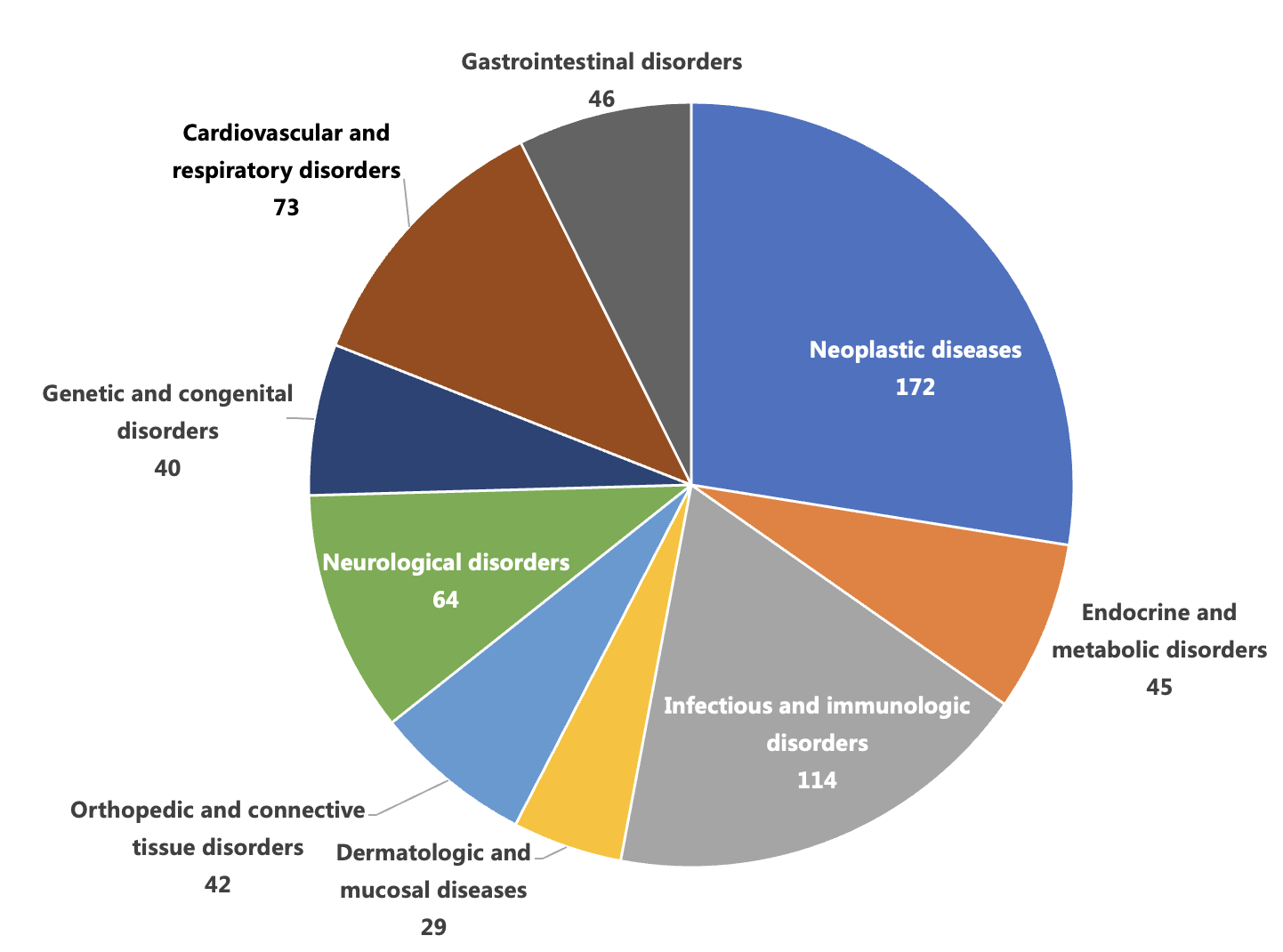}}
    \caption{Distributions of image modalities (a) and diagnostic categories (b) in \emph{MedReaMM}, reflecting the benchmark’s clinical realism and coverage.}
    \label{fig:modality_diagnosis_pie}
\end{figure*}

\begin{figure*}[!h]
    \centering
    \subfigure[Textual Inputs]{\label{fig:wordcloud_input}\includegraphics[width=0.4\textwidth]{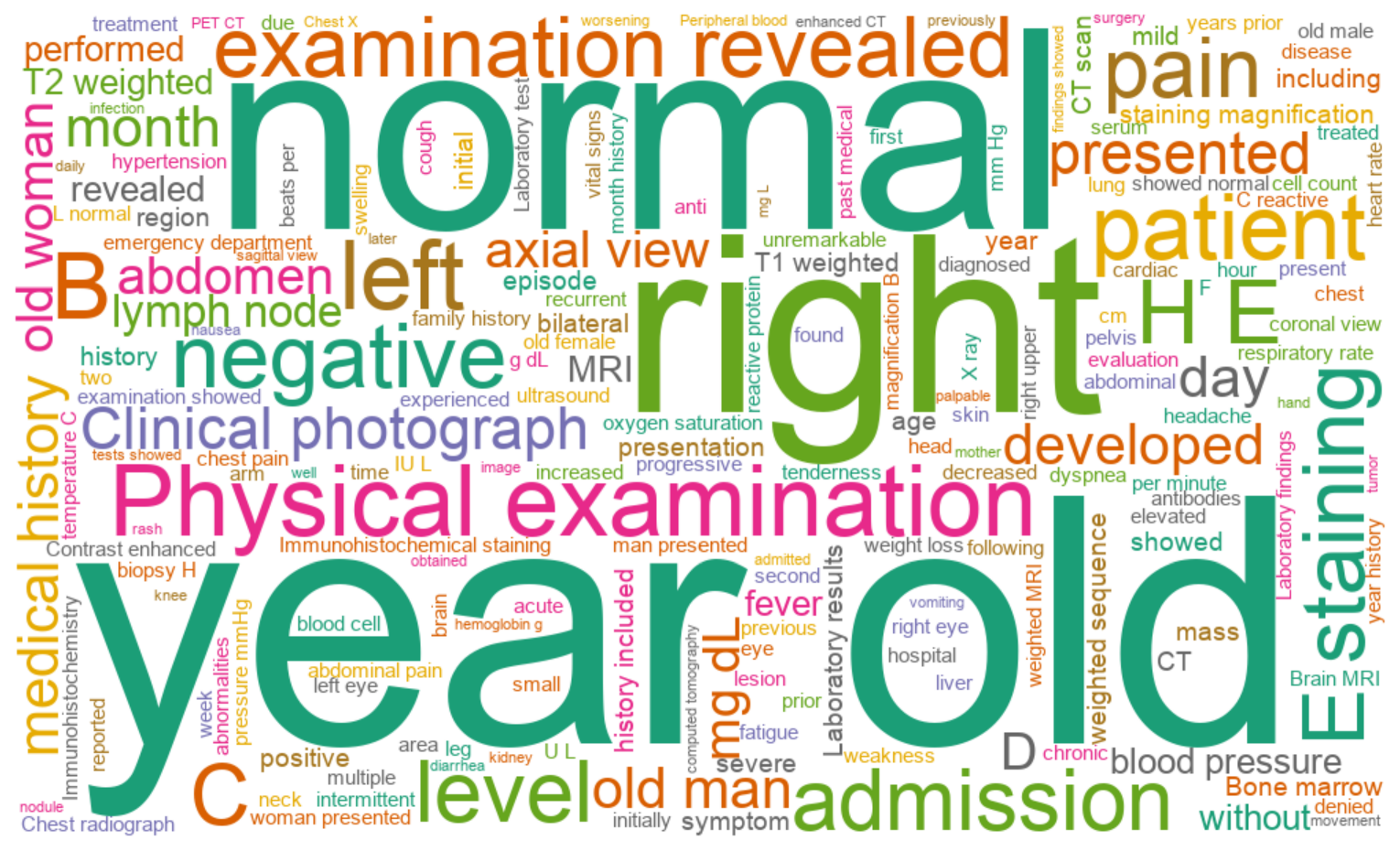}}
    \hfill
    \subfigure[Diagnosis Terms]{\label{fig:wordcloud_diagnosis}\includegraphics[width=0.4\textwidth]{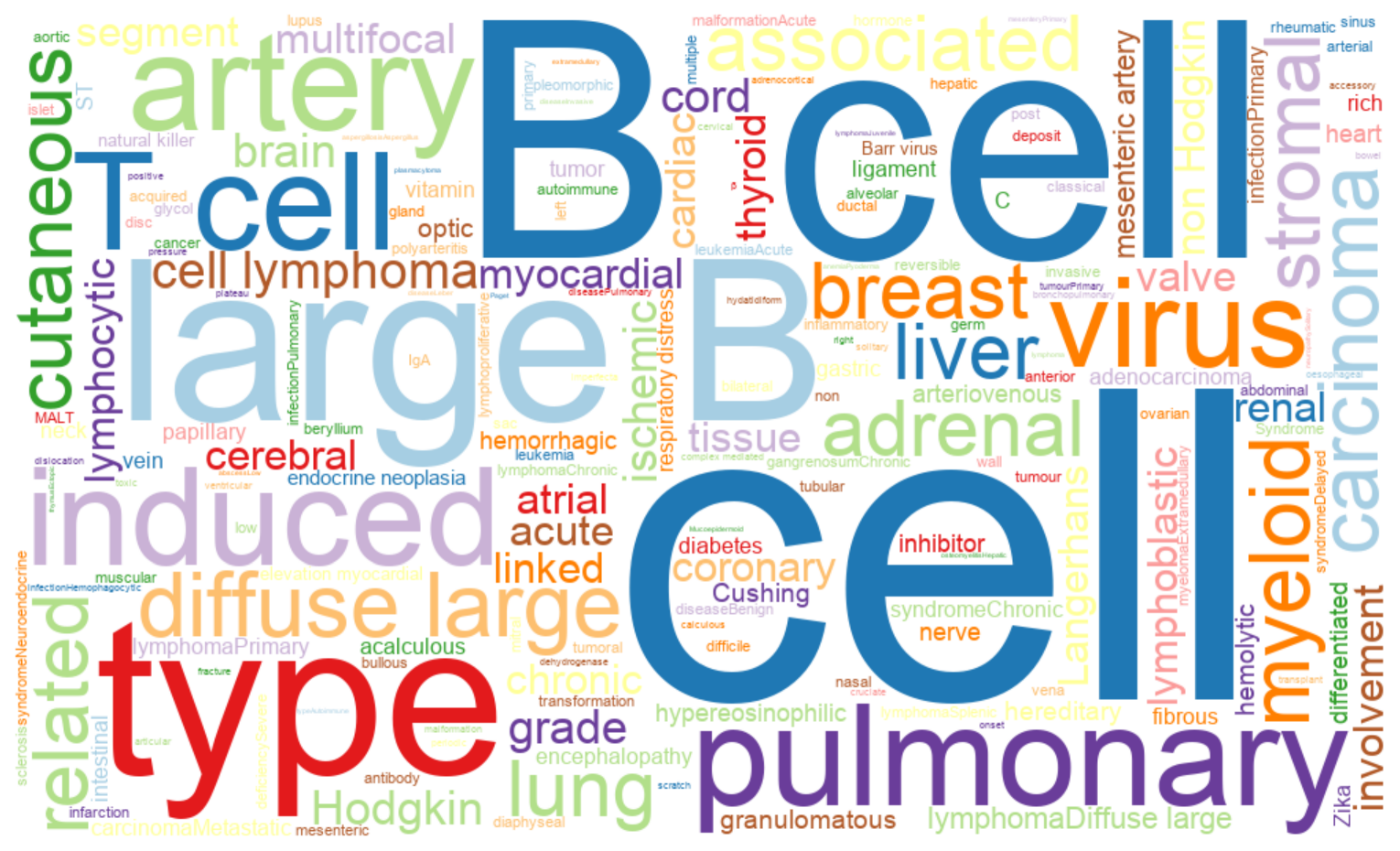}}
    \caption{Word clouds of textual inputs (a) and diagnosis terms (b) in \emph{MedReaMM}.}
    \label{fig:wordclouds}
\end{figure*}

\subsection{Image Modalities and Diagnostic Categories}

To obtain fine-grained image-type annotations, we utilized GPT-4.1 \cite{achiam2023gpt} in combination with the detailed captions ($C_d$) associated with each image. 
This enabled us to categorize images into 15 distinct types, 13 of which correspond to standard medical imaging modalities, including whole slide images (WSI), CT, MRI, ultrasound, endoscopy, and PET.

Similarly, to categorize disease types, we provided GPT-4.1 with the final diagnoses ($D$) from all cases and instructed it to generate case-level disease category labels.
This automated classification enabled a comprehensive mapping of diagnostic coverage across organ systems.

Figure~\ref{fig:modality_diagnosis_pie} presents the distribution of image modalities (left) and disease categories (right), reflecting the benchmark's multimodal and multi-organ characteristics.

\begin{table}[!h]
\centering
\begin{tabular}{cc|cc}
\toprule
\makecell{\textbf{\# Images} \\ \textbf{per Case}} & \textbf{\# Cases} & \makecell{\textbf{\# Diagnoses} \\ \textbf{per Case}} & \# Cases\\
\midrule
1  & 131 & 1 & 334 \\
2  & 165 & 2 & 197 \\
3  & 151 & 3 & 67  \\
4  & 100 & 4 & 21  \\
5-10  & 78  & 5 & 6   \\
\bottomrule
\end{tabular}
\caption{Distributions of the number of images (left) and diagnoses (right) per case in \emph{MedReaMM}. Many cases contain multiple images and diagnoses, reflecting the complexity of real-world clinical data.}
\label{tab:quantity_stats}
\end{table}

\subsection{Image and Diagnosis Quantity Statistics}

Beyond categorical composition, we analyze the number of images and diagnoses per case, which are critical for understanding the complexity of diagnostic tasks in the benchmark.

Table~\ref{tab:quantity_stats} summarizes the distributions of image count and diagnosis count per case. Notably, a large proportion of cases include multiple images, emphasizing the dataset’s departure from conventional VQA-style benchmarks that rely primarily on single-image inputs \cite{he2020pathvqa,liu2021slake,lau2018dataset}. Likewise, \emph{MedReaMM} supports multi-diagnosis supervision: 79\% of cases have two or more diagnostic labels, aligning with the clinical reality of comorbid conditions.

In total, the dataset comprises 1043 diagnostic terms.

\begin{algorithm}[t]
\caption{ICD Encoding Procedure}
\label{alg:icd_encoding}
\begin{algorithmic}[1]
\REQUIRE Search text $s$
\ENSURE ICD code $c$ (or \texttt{None} if no match)
\STATE $c \leftarrow$ \texttt{auto\_code}($s$)
\IF{$c \neq$ \texttt{None}}
    \RETURN $c$
\ENDIF
\STATE $c \leftarrow$ \texttt{flex\_search}($s$)
\IF{$c \neq$ \texttt{None}}
    \RETURN $c$
\ENDIF
\RETURN \texttt{None}
\end{algorithmic}
\end{algorithm}

\subsection{Vocabulary Distribution Visualization}

To further illustrate the diversity of diagnostic concepts and textual descriptions, we generate word clouds summarizing both the textual inputs and the diagnostic terms.

Figure~\ref{fig:wordclouds} shows two word clouds: the left summarizes the frequency of clinical terms in the structured inputs, and the right highlights the distribution of diagnosis terms across all cases.
These visualizations reveal the lexical richness and diagnostic variability captured in \emph{MedReaMM}.

\section{Dataset Scope and Limitations}
\label{appendix:data_scope}
\subsection{Expert-Level Case Selection}
\emph{MedReaMM} focuses on high-level clinical reasoning rather than routine primary care screening. The dataset primarily features cases characterized by diagnostic complexity, such as rare diseases and atypical multi-system pathologies. This focus establishes \emph{MedReaMM} as a benchmark for evaluating AI performance in "long-tail" clinical scenarios where human experts typically require multidisciplinary consultation.

\subsection{Information Density and Narrative Structure}
The cases in \emph{MedReaMM} provide expert-curated narratives that represent the distilled cognitive output of clinicians. This high-density information format serves to evaluate the upper limits of a model's logical synthesis and evidence integration. Assessing performance on these standardized yet complex inputs is a foundational step in establishing a reasoning baseline prior to the evaluation of noisier, fragmented real-world clinical data.

\subsection{Data Distribution and Scale}
The benchmark consists of 625 expert-validated cases with ICD-11 standardized annotations. The distribution is naturally concentrated in fields with high diagnostic uncertainty, such as internal medicine and oncology. This composition is a deliberate structural choice to stress-test LMMs under expert-level requirements. The priority of \emph{MedReaMM} is the depth of multimodal evidence per case rather than the breadth of common clinical presentations.

\subsection{Clinical Decision Support Alignment}
\label{appendix:metrics_alignment}
The evaluation metrics, specifically Top-5 and Top-10 recall, align with the clinical utility of AI as a decision-support tool. In complex diagnostic dilemmas, the primary function of such systems is to provide a broad range of differential diagnoses for clinician review. The benchmark measures the model's capacity to identify correct diagnostic leads, assisting in cases of diagnostic uncertainty.

\section{ICD Encoding}
\label{appendix:icd}

The \emph{International Classification of Diseases} (ICD), developed by the World Health Organization (WHO), is the most widely adopted global standard for medical diagnostic coding \cite{o2005measuring}. 
ICD-11 \cite{harrison2021icd} is the latest and most comprehensive version of the classification system, designed to facilitate consistent diagnosis and health condition reporting across international healthcare systems. 
The WHO provides an official ICD API to support automated query and encoding services, including two key functionalities: \texttt{auto-code} and \texttt{flexible search}.

The \texttt{auto-code} function takes a search string as input and returns the best-matched ICD-11 code. 
In contrast, the \texttt{flexible search} function attempts to identify the closest ICD-11 match even if parts of the input text are unmatched. 
According to official recommendations, \texttt{flexible search} should only be used when \texttt{auto-code} fails to return a result.

During the construction of \emph{MedReaMM}, we exclusively used the \texttt{auto-code} function, supplemented with manual verification and correction, to ensure the accuracy and validity of the ground-truth diagnostic labels.

In the evaluation phase, for computing ICD-based matching metrics, we applied a hierarchical ICD encoding procedure: the system first attempts \texttt{auto-code}; if unsuccessful, it falls back to \texttt{flexible search}.
If both methods fail to yield a valid code, the diagnosis is considered unmatched. The process is summarized in Algorithm~\ref{alg:icd_encoding}.
\begin{figure*}[!h]
    \centering
    \includegraphics[width=\linewidth]{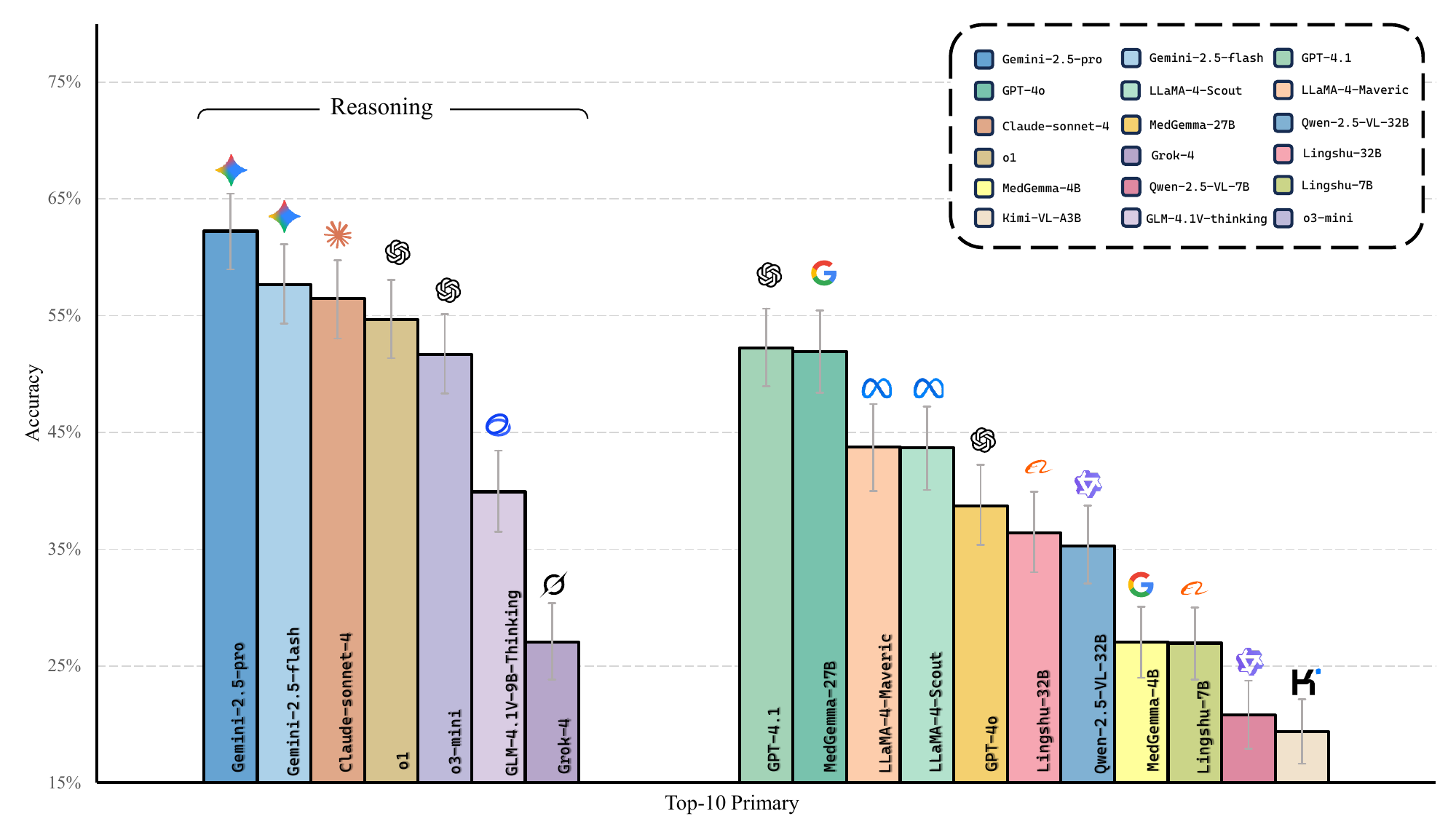}
    \caption{Bootstrap-estimated 95\% confidence intervals for \emph{Primary Diagnosis Recall}.}
    \label{fig:ci}
\end{figure*}

\section{Human vs. Model Evaluation}
\label{appendix:human_vs_lmm}
\begin{table}[!h]
\centering

\begin{tabular}{lcc}
\toprule
\textbf{Method} & \textbf{Accuracy (\%)} & \textbf{DSC} \\
\midrule
Human & 10.87 & 0.2201 \\
\midrule
Gemini-2.5-Flash & 43.48 & 0.6203 \\
GPT-4.1 & 47.83 & 0.6453 \\
\bottomrule
\end{tabular}

\caption{Comparison of diagnostic accuracy between human gastroenterologists and LMMs on gastrointestinal disorders.}
\label{tab:human_vs_llm}
\end{table}

To evaluate the performance gap between LMMs and human expertise, we conducted a controlled diagnostic study using a specialized subset of gastrointestinal cases from \emph{MedReaMM}.
Specifically, these cases were reformulated into a multiple-choice format to allow for standardized comparison. For each case, the ground-truth diagnosis was paired with distractors sampled from the incorrect predictions generated by models in our main experiment. This setup resulted in 46 questions, consisting of 27 single-answer and 19 multiple-answer questions.

In clinical practice, interpretation of medical images often involves collaboration with radiologists or other imaging specialists.
To decouple diagnostic reasoning from image interpretation expertise, human clinicians are provided with the original detailed image captions ($C_d$), which contain expert-level interpretations of the associated medical images.
The same inputs are provided to the evaluated models to ensure consistency between human and model evaluation settings.

Three independent gastroenterologists completed the task, and their performance was assessed using two complementary scoring methods.
First, we adopt a strict \textbf{Accuracy} metric, where a response is considered correct only if it exactly matches the ground truth—requiring all correct options to be identified in multiple-answer questions.
Second, to capture a more nuanced view of diagnostic coverage, we employ the \textbf{Sorensen-Dice Coefficient} ($DSC$) to measure the overlap between the predicted diagnostic sets and the ground truth. 
The coefficient is defined as:
\begin{equation}
    DSC = \frac{2 |X \cap Y|}{|X| + |Y|}
\end{equation}
where $X$ and $Y$ represent the predicted and reference sets, respectively.
As shown in Table \ref{tab:human_vs_llm}, while the human experts achieved an accuracy of 10.87\% and a $DSC$ of 0.2201, the AI models demonstrate a more robust capacity for identifying relevant diagnostic components within these complex clinical scenarios.

\noindent \textbf{Analysis of the Human-Model Gap}\quad The observed performance disparity between human clinicians and LMMs can be primarily attributed to the extreme complexity and interdisciplinary nature of the cases in \emph{MedReaMM}.

First, there is a fundamental difference between individual capacity and collective expertise. The cases in our benchmark, predominantly sourced from top-tier journals such as the \emph{NEJM}, typically represent "diagnostic dilemmas" that require a Multi-Disciplinary Team (MDT) for resolution in real-world practice. In contrast, our human baseline reflects the performance of individual specialists working without peer consultation or external search tools.

Second, qualitative feedback from the participating specialists indicated that many cases involved rare diseases or multi-system complications that fall outside their specific sub-specialties. For a single expert, identifying every granular diagnostic component in such high-stakes, low-frequency scenarios is inherently challenging. These results underscore that while individual human experts are constrained by specialized knowledge boundaries, LMMs leverage a broader knowledge base and robust multimodal integration to serve as a powerful adjunct in managing complex, long-tail clinical cases.

\section{Robustness Analysis}
\label{appendix:robust}

\subsection{Results With Confidence Intervals}

To better compare diagnostic performance across models, we computed 95\% confidence intervals for selected models on the Top-10 \emph{Primary Diagnosis Recall} task using the bootstrap method \cite{diciccio1996bootstrap} with 5000 resamples, as shown in Figure~\ref{fig:ci}.

From the results, we observe a stepwise distribution of model performance: models form distinct tiers in terms of primary diagnosis recall. 
In some cases, confidence intervals between adjacent models overlap, indicating that the observed differences may not be statistically significant. 
In contrast, non-overlapping intervals suggest more robust and reliable distinctions in diagnostic capabilities. 

\subsection{Prompt Robustness in Evaluation}

Inspired by prior studies \cite{ghazal2013bigbench}, we conducted a multi-prompt experiment to assess the robustness of our evaluation. Specifically, we designed three semantically equivalent but syntactically varied prompts and evaluated model performance under each. Prompt 1 is the prompt used in main evaluation and prompt 2, prompt 3 are shown in the \emph{Prompts} section in the Appendix (Figure \ref{fig:prompt_2} and \ref{fig:prompt_3}).

Across most tested models, the performance fluctuation remained within a 3.3\% margin for key metrics such as Top-5 Primary and Complete Diagnosis Recall. 
This suggests that our evaluation results are stable and not significantly affected by minor variations in prompt wording.

\begin{table}[htbp]
  \centering
  \resizebox{0.45\textwidth}{!}{%
  \begin{tabular}{l c c c}
  \toprule
  \multirow{2}{*}[-1ex]{\textbf{Model}} & \multirow{2}{*}[-1ex]{\textbf{Prompt Type}} & \multicolumn{2}{c}
  {\textbf{Top-5}} \\
  \cmidrule(lr){3-4}
  & & \textbf{Primary} & \textbf{Complete} \\
  \midrule
  \multirow{3}{*}{GPT-4o} & prompt1 & 48.9 & 25.0 \\
  & prompt2 & 49.2 & 26.7 \\
  & prompt3 & 49.8 & 28.7 \\
  \midrule
  \multirow{3}{*}{Claude-Sonnet-4} & prompt1 & 51.4 & 34.4 \\
  & prompt2 & 49.0 & 34.0 \\
  & prompt3 & 48.3 & 34.2 \\
  \midrule
  \multirow{3}{*}{Gemini-2.5-Flash} & prompt1 & 55.0 & 38.6 \\
  & prompt2 & 53.3 & 40.0 \\
  & prompt3 & 54.3 & 38.8 \\
  \midrule
  \multirow{3}{*}{Qwen2.5-VL-72B} & prompt1 & 38.9 & 25.0 \\
  & prompt2 & 39.0 & 25.7 \\
  & prompt3 & 38.3 & 25.8 \\
  \bottomrule
  \end{tabular}
  }
  \caption{Performance Comparison by Model and Prompt Type (Top-5) on the 60-case subset.}
  \label{tab:top5_prompt_performance}
  \end{table}

\section{Supplementary Experiment Details}

\subsection{Data Leakage}

As pretraining corpora expand, LMMs may memorize specific cases and answer by reproducing seen content. 
To examine potential data leakage, we analyzed model performance by case report year, assuming 2025 cases were not in the training data. 
In Figure \ref{fig:year}, We observed no notable performance drop on 2025 cases, suggesting little to no leakage or negligible impact on evaluation.

\subsection{Performance Across Diagnostic Categories}
\begin{figure*}[!h]
    \centering
    \includegraphics[width=\linewidth]{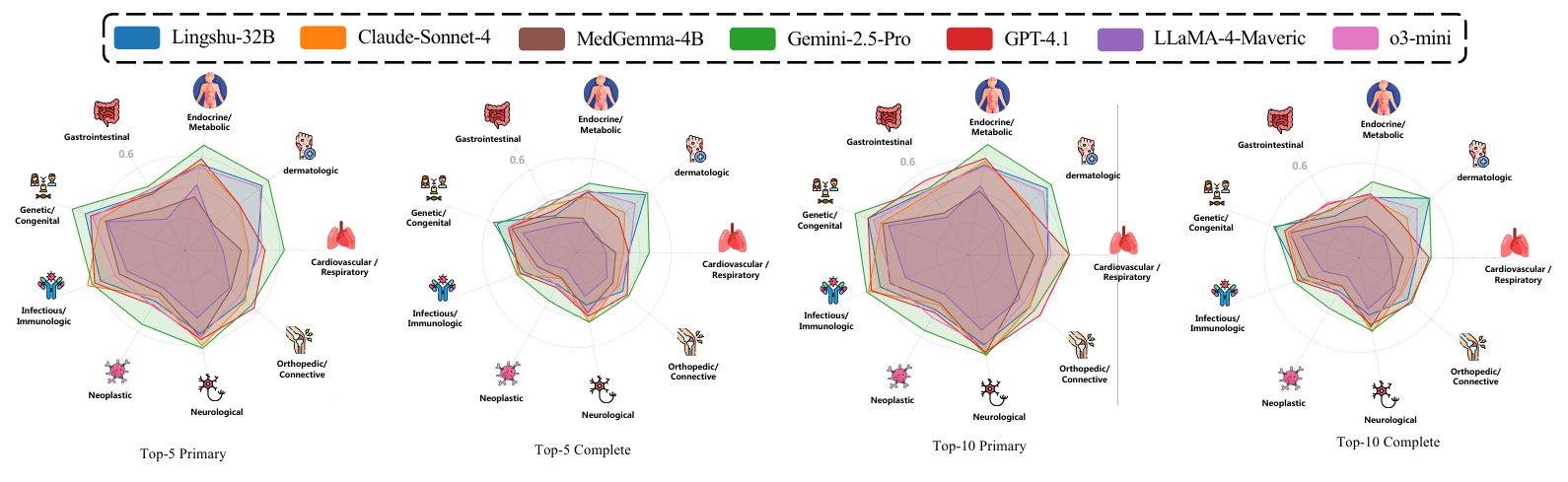}
    \caption{Model performance across disease categories under different evaluation settings.}
    \label{fig:radar_category}
\end{figure*}

To further examine model robustness and diagnostic generalization, we analyzed performance across different disease categories under four evaluation settings: Top-5 Primary, Top-5 Complete, Top-10 Primary, and Top-10 Complete Diagnosis Recall. All test cases were grouped using GPT-4.1, including cardiovascular, dermatologic, endocrine, gastrointestinal, congenital, infectious, neoplastic, neurological, and orthopedic disorders.

\begin{figure}
    \centering
    \includegraphics[width=\linewidth]{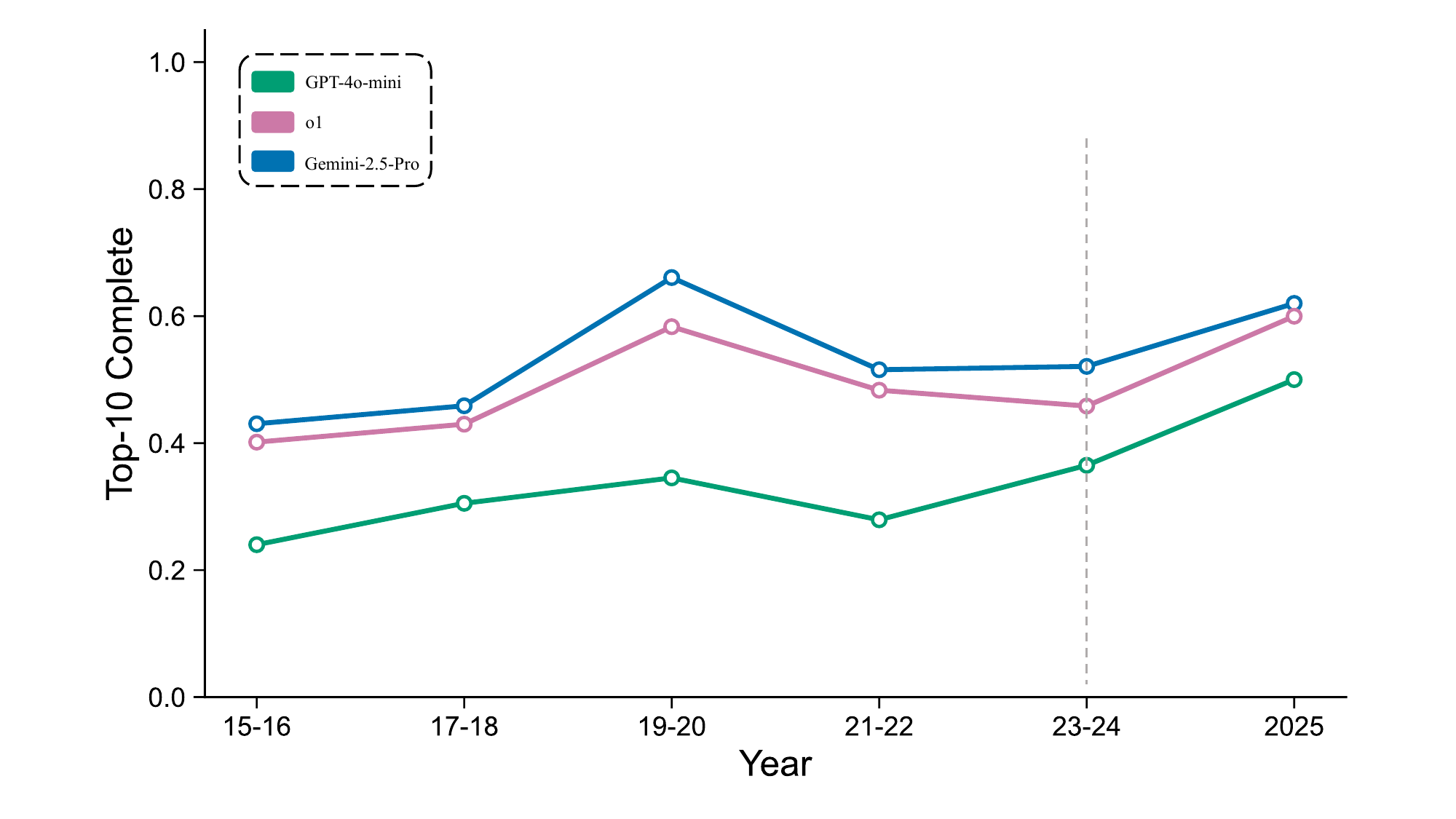}
    \caption{Model performance by publication year of case reports. No significant drop is observed for 2025 cases, suggesting minimal data leakage from pretraining corpora.}
    \label{fig:year}
\end{figure}

Figure~\ref{fig:radar_category} presents radar charts showing category-wise performance for representative models. 
Across all settings, models tend to perform consistently better on more common and visually distinct disease types, such as \textbf{endocrine/metabolic} and \textbf{gastrointestinal} disorders. In contrast, performance is generally lower on \textbf{neoplastic} and \textbf{neurological} cases, which often require more nuanced clinical reasoning or involve subtle imaging features.

Among the models, Gemini-2.5-Pro\cite{comanici2025gemini} shows relatively balanced performance across categories and settings, while smaller models such as MedGemma-4B \cite{sellergren2025medgemma} and Lingshu-32B \cite{xu2025lingshu} display more pronounced category-specific weaknesses. 
Notably, performance gaps between models narrow as the recall setting becomes more lenient, but category-level discrepancies persist.

\section{Human Annotation}
\label{appendix:human_anno}

During the construction of \emph{MedReaMM}, we employed human annotators in two key processes: \emph{label leakage detection} and \emph{clinical expert review}. The details of each step are described below.

\noindent\textbf{Label Leakage Detection} \quad Three second-year medical graduate students were recruited to examine and validate the data. 
All annotators had foundational medical knowledge and the ability to read and understand case reports. 
For each case, we provided them with the Medical Records ($R$), Medical Images and their corresponding Captions ($\{(I_j, C_j)\}_{j=1}^n$), the final Diagnosis ($D$), and the original source case report.

The annotators were asked to examine the following aspects:  
1) Whether the primary diagnosis appears in the medical records or image captions;  
2) Whether any caption includes clinical findings or subjective interpretations;  
3) Whether the image itself contains post-diagnostic information (e.g., recovery status).

Given their training level, we did not consider the student annotators qualified to revise the provided medical records or captions in cases where label leakage was identified. 
Therefore, for criteria (1) and (2), if at least two annotators judged that label leakage existed in the text information, the corresponding case was removed. 
For criterion (3), if two or more annotators agreed that an image contained inappropriate information, the image was either excluded or the violating region was cropped.

Over the course of approximately 45 days, the annotators reviewed 1527 cases extracted through the multimodal data pipeline and retained 1193 cases for the next stage. 

\paragraph{Clinical Expert Review}  
In the final phase of dataset construction, we engaged two senior clinicians to perform professional validation of the dataset. For each case, they were provided with the Medical Records ($R$), Medical Images and their corresponding Captions ($\{(I_j, C_j)\}_{j=1}^n$), the final Diagnosis ($D$), and the original source case report.

\begin{figure}[!h]
    \centering
    \subfigure[]{\includegraphics[width=0.23\textwidth]{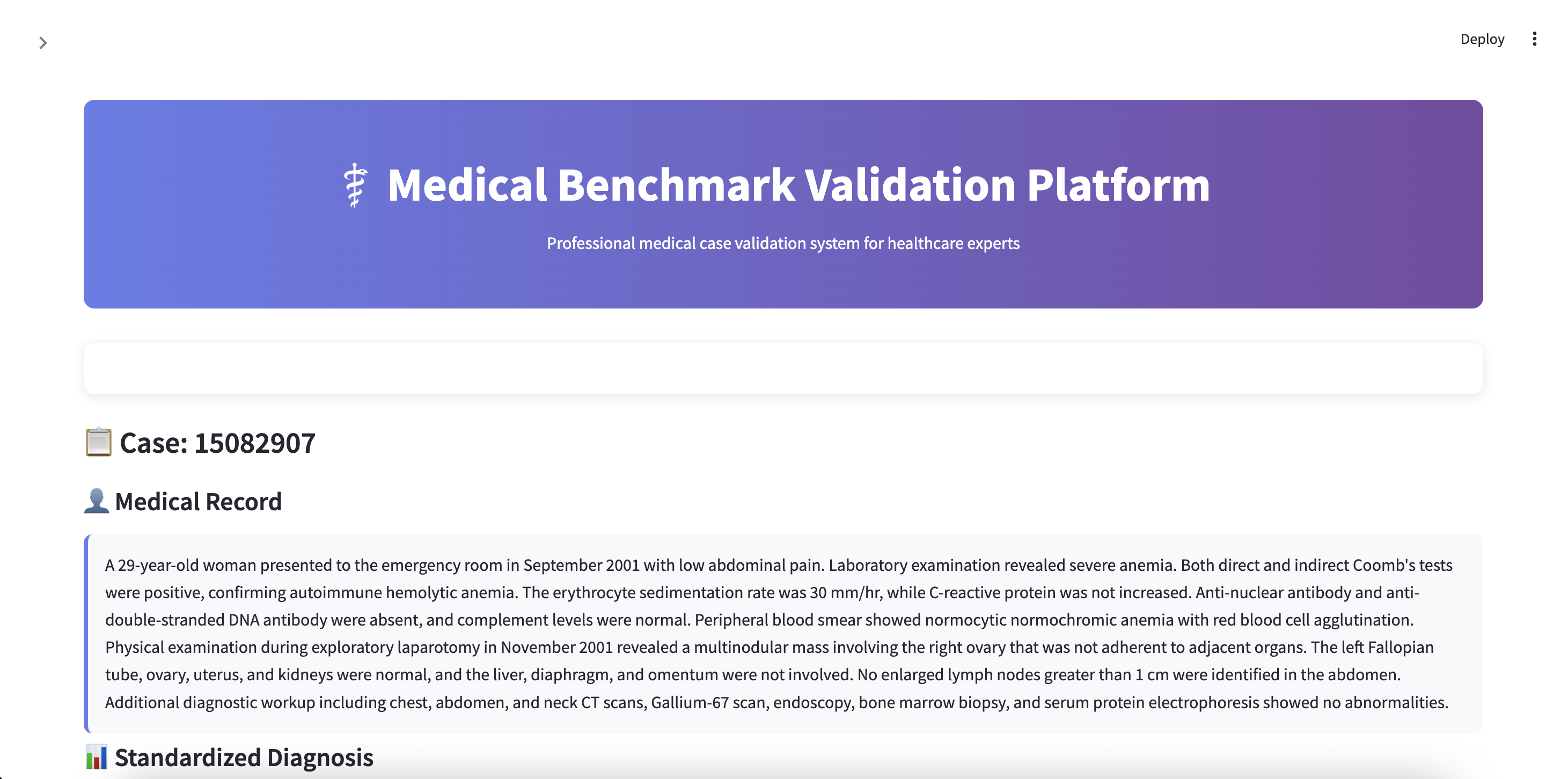}}
    \subfigure[]{\includegraphics[width=0.23\textwidth]{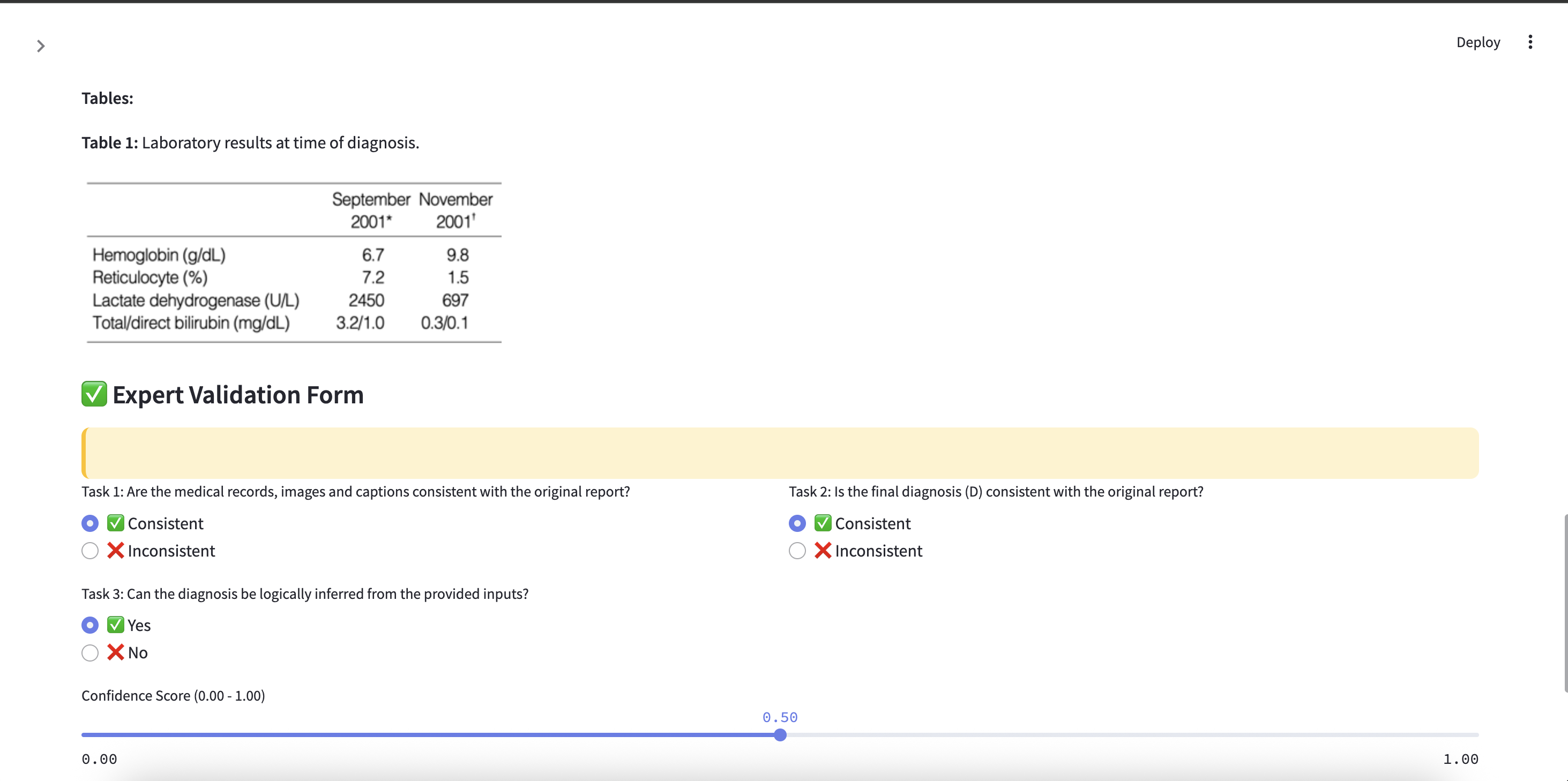}}
    \caption{The UI interface used in the clinical expert validation phase.}
    \label{fig:expert_review_ui}
\end{figure}

The experts were asked to verify: (1) whether the provided inputs ($R$, $\{(I_j, C_j)\}$) and the final diagnosis ($D$) were consistent with the content of the original report, and (2) whether the diagnosis could be logically inferred from the provided input. 
To facilitate this process, we designed a structured annotation interface for the experts, displaying each case with rating options and a confidence score field for their assessment (see Figure~\ref{fig:expert_review_ui}). 
Following the expert assessment, 625 cases were retained as the final version of \emph{MedReaMM}.

Table~\ref{tab:case_selection_summary} summarizes the number of remaining cases after each data processing stage. 
Starting from 2867 case reports, we excluded those lacking images or focused solely on etiology, yielding 1527 cases. 
Label leakage detection reduced this to 1193, all of which were retained after diagnosis encoding. 
Finally, clinical experts removed cases where the diagnosis could not be fully inferred from the input, resulting in 625 high-quality cases. 
Although strict inference criteria led to substantial reduction, we note that incomplete inputs are common and still meaningful in real-world clinical practice.

\begin{table}[h]
\centering
\resizebox{0.45\textwidth}{!}{%
\begin{tabular}{lc}
\toprule
\textbf{Stage} & \textbf{Remaining Cases}  \\
\midrule
Case Report Collection & 2867  \\
Multimodal Data Extraction & 1527 \\
Label Leakage Detection & 1193 \\
Diagnosis Encoding & 1193 \\
Clinical Expert Review & 625\\
\bottomrule
\end{tabular}
}
\caption{
Case selection process across five stages.}
\label{tab:case_selection_summary}
\end{table}

\section{Human Evaluation}
\label{appendix:human_eval}

To better understand the factors influencing model diagnostic capabilities, we evaluated model outputs across seven diagnostic dimensions using GPT-5. To assess the validity of GPT-5's evaluations, we randomly sampled a subset of examples and asked human annotators to rate them using the same criteria.

Three senior-year medical students independently rated each of the seven dimensions. We report both the agreement between the human scores and those from GPT-5, as well as the inter-annotator agreement among the three medical students. The results are summarized in Table~\ref{tab:human_eval_agreement}.

\begin{table}[!h]
\centering
\resizebox{0.45\textwidth}{!}{%
\begin{tabular}{lccc}
\toprule
\textbf{Evaluation Dimension} & \makecell{\textbf{Sample}\\\textbf{Size}} & \textbf{GPT-H} & \textbf{H-H} \\
\midrule
\emph{Causes}  & \multirow{4}{*}{40} & 0.93 & 0.88    \\
\emph{Symptoms} &  & 0.93  & 0.90  \\
\emph{Diagnostic Tests} &   & 0.88   & 0.85 \\
\emph{Treatments}  &    & 0.90 & 0.85 \\
\midrule
\emph{Key Feature Detection}  & \multirow{2}{*}{70} & 0.94  & 0.91      \\
\emph{Medical Interpretation} & & 0.91    & 0.89 \\
\midrule
\emph{Reasoning Path Accuracy}& 25  & 0.88  & 0.84   \\
\bottomrule
\end{tabular}
}
\caption{Agreement between GPT-5 and human annotators, and inter-annotator agreement among three medical students. GPT-H means GPT-Human agreement; H-H means Human-Human agreement.}
\label{tab:human_eval_agreement}
\end{table}

\section{Prompts}
\label{appendix:prompts}
This section summarizes all prompts used during dataset construction and evaluation.

\noindent\textbf{Dataset Construction:} We used Claude Sonnet 4 to extract structured information from PDFs, guided by the prompt in Figure~\ref{fig:prompt_content_extraction}. Its native PDF support allowed direct attachment without preprocessing. To generate final captions ($C$), we employed GPT-5 with the prompt in Figure~\ref{fig:prompt_remove}, removing clinical findings from the detailed caption ($C_d$) and retaining only image metadata. And we use prompt in Figure \ref{fig:prompt_reasoning_path_extraction_gt} to extract reasoning path from case reports.

\noindent\textbf{Evaluation:} The main evaluation prompt is shown in Figure~\ref{fig:prompt_main}, also used as Prompt 1 in our robustness study. Prompt 2 and Prompt 3 appear in Figures~\ref{fig:prompt_2} and~\ref{fig:prompt_3}, respectively.

\noindent\textbf{Analysis:} To examine factors influencing diagnostic performance, we used prompts in Figures~\ref{fig:prompt_disease_kowledge}, \ref{fig:prompt_image_understanding}, and~\ref{fig:prompt_reasoning_path_extraction} to assess disease knowledge, image understanding, and clinical reasoning. Corresponding evaluation prompts are shown in Figures~\ref{fig:prompt_disease_eval}, \ref{fig:prompt_image_understanding_eval} and \ref{fig:prompt_reasoning_path_eval}.

\section{Case Study}
\label{appendix:case}

The case shown in Figure \ref{fig:case} involves a 35-year-old female patient with a four-year history of recurrent pneumonia, who was ultimately diagnosed with intralobar pulmonary sequestration complicated by actinomycosis infection. The diagnosis was based on chest CT findings revealing a lesion in the left lower lobe and histopathological examination confirming characteristic actinomyces colonies.

Although stronger models such as Gemini-2.5-Pro correctly identified the primary condition—pulmonary sequestration—all models failed to recognize the complication of actinomycosis. 
This limitation may stem from two underlying factors. 
First, models tend to focus on the most prominent and easily inferred diagnoses while overlooking coexisting conditions that are less obvious but clinically significant. 
Second, current models may still struggle to extract fine-grained pathological details from medical images, limiting their ability to incorporate such findings into diagnostic reasoning. 
Together, these challenges highlight a broader limitation in the integration of multimodal clinical signals.

\onecolumn
\begin{figure}
    \centering
    \resizebox{0.9\textwidth}{!}{%
    \begin{tcolorbox}[title=Case Study: Intralobar Pulmonary Sequestration with Actinomycosis (part 1), colback=white, colframe=orange, width=\textwidth, breakable]

\begin{minipage}[t]{\textwidth}
\textbf{Medical Record:} A 35-year-old female patient attended the outpatient clinics of the Thoracic Surgery Service, referred by her general practitioner for dyspnea and persistent episodes of coughing. Physical examination revealed bilateral diffuse crackles upon auscultation of the lungs. She had a 4-year history of recurrent episodes of pneumonia, which had required hospitalization and antibiotic management. There was no past history of cigarette smoking or secondhand smoke exposure and her family history was negative.
\end{minipage}

\begin{figure}[H]
    \centering
    \subfigure[(A) Histopathological examination, H\&E staining, 10× magnification. (B) Histopathological examination, H\&E staining, 100× magnification. Gram staining.]{
        \includegraphics[width=0.9\linewidth]{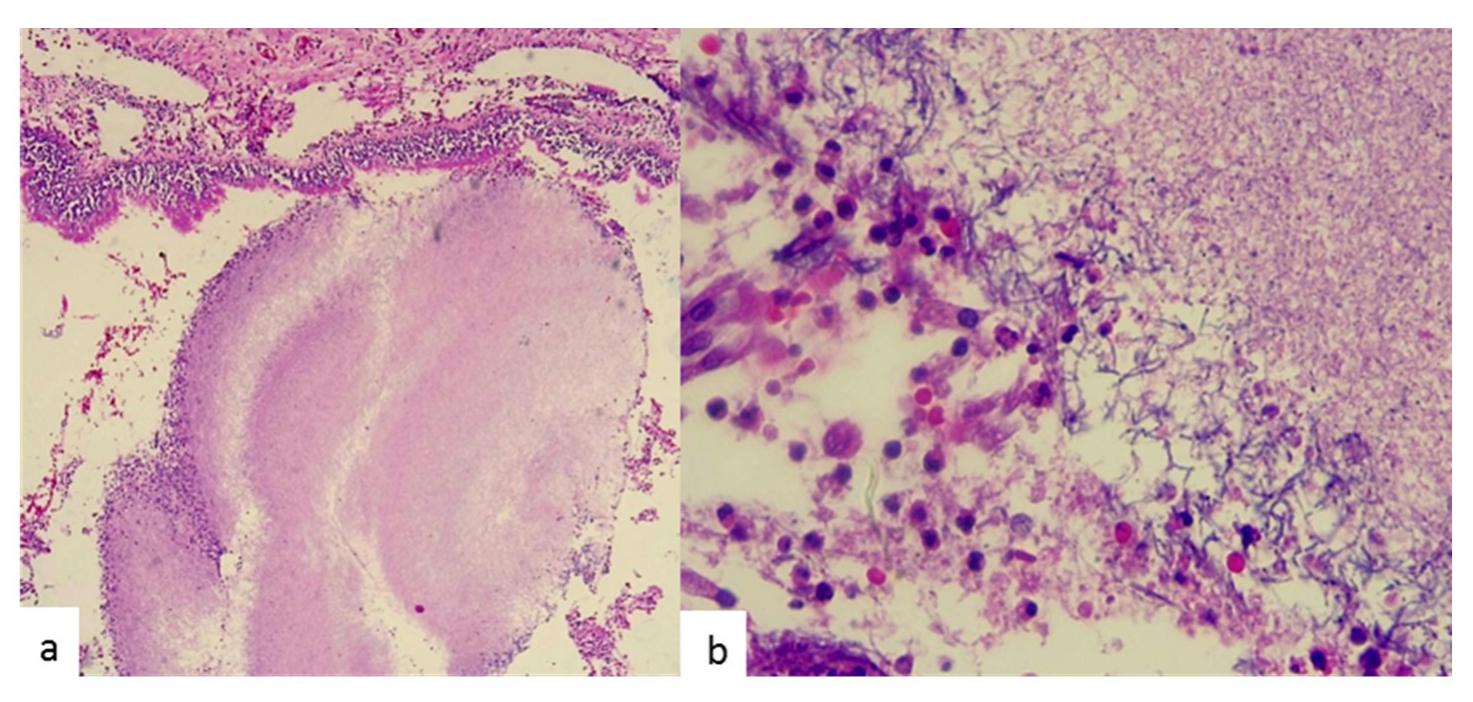}}
        
    \subfigure[Laboratory test results on admission.]{
        \includegraphics[width=0.45\linewidth]{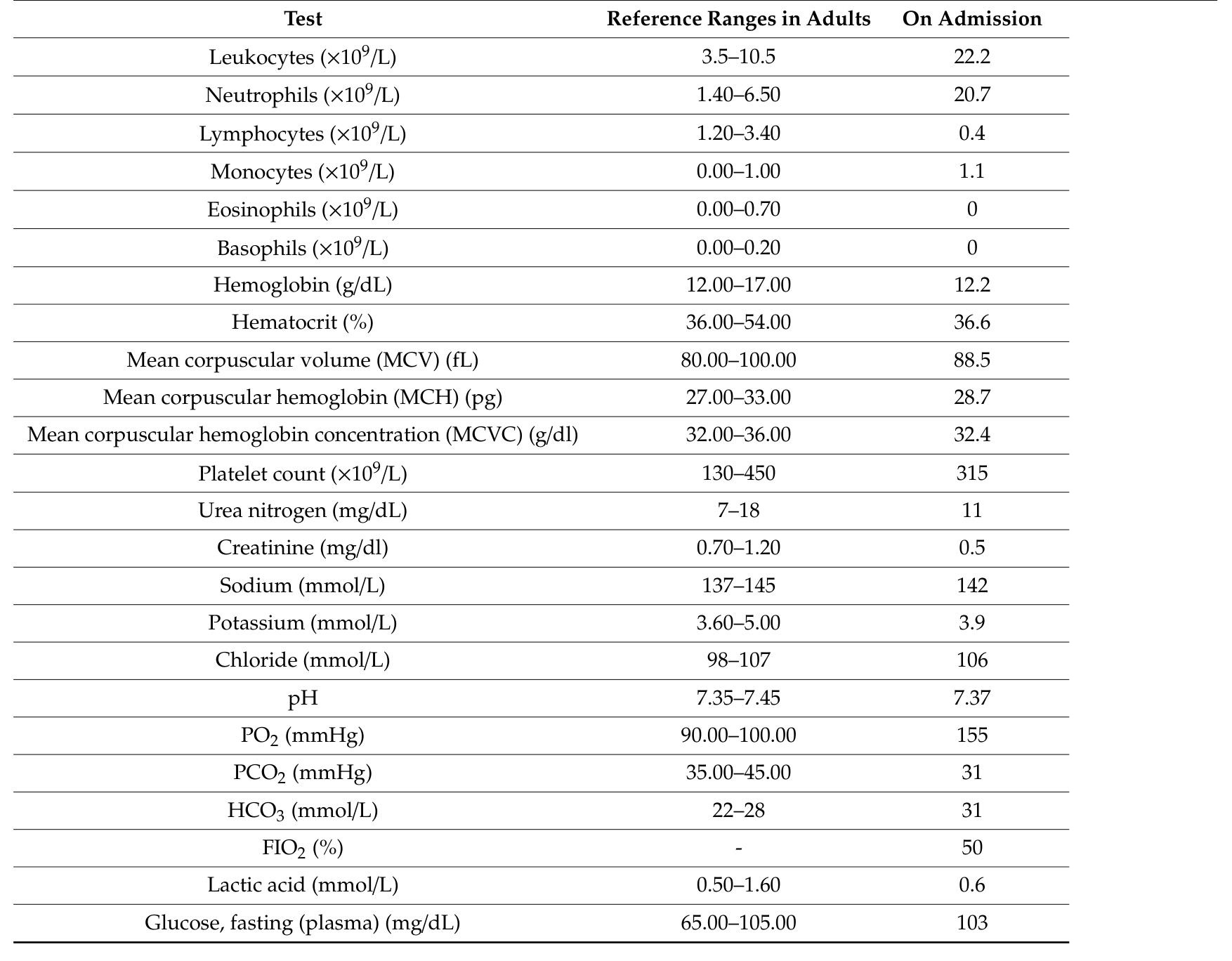}}
    \subfigure[(A, C) Chest CT with contrast, axial view, posterior segment of the left lower lobe. (B, D) Chest CT with contrast, coronal view, posterior segment of the left lower lobe.]{
        \includegraphics[width=0.45\linewidth]{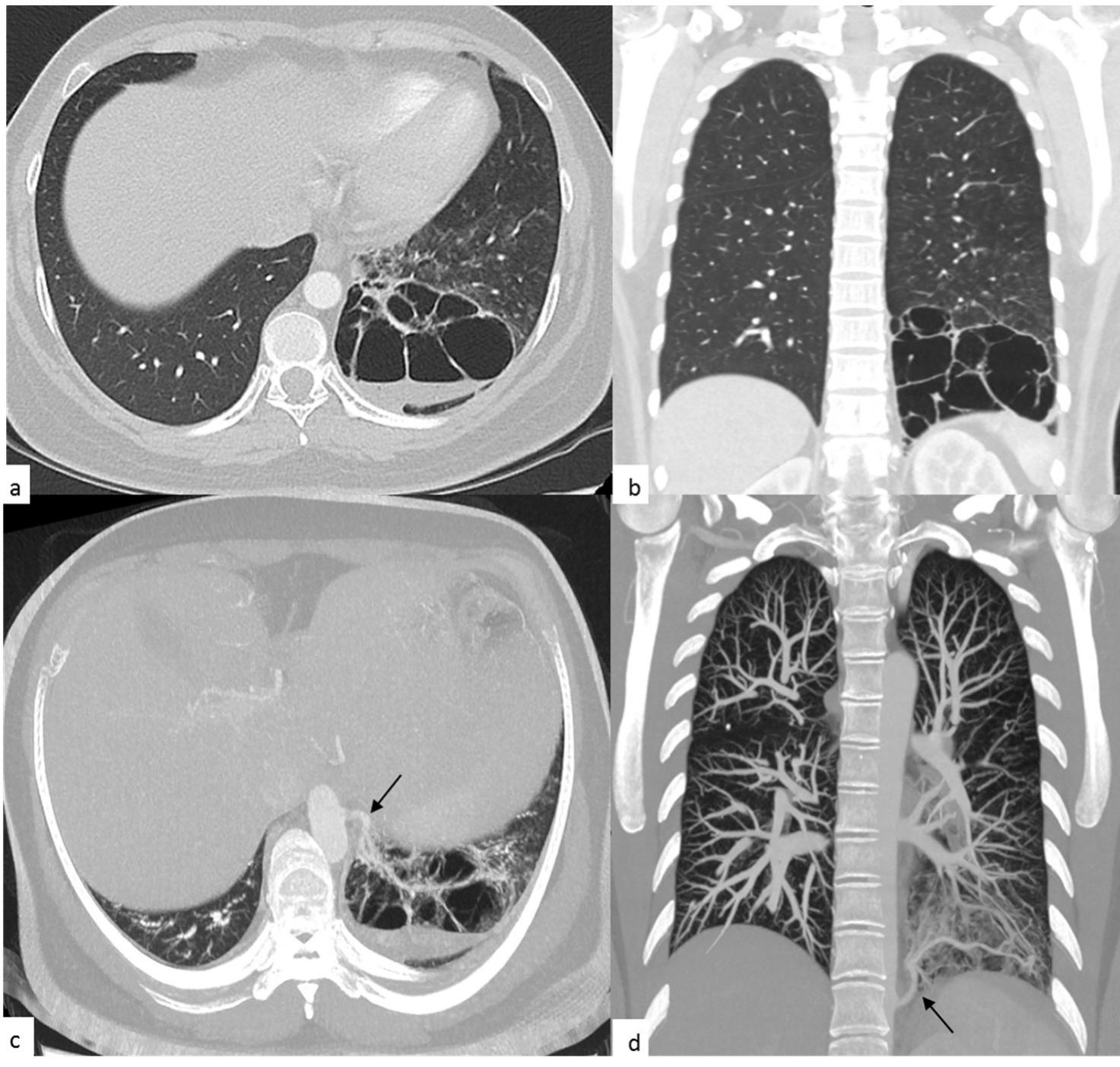}}
\end{figure}

\textbf{Diagnosis:}
\begin{itemize}
    \item Intralobar bronchopulmonary sequestration \emph{(LA75.6)}
    \item Actinomyces infection \emph{(1C10.Z)}
\end{itemize}

\end{tcolorbox}
}
\end{figure}

\begin{figure}
    \centering
    \resizebox{0.8\textwidth}{!}{%
    \begin{tcolorbox}[title=Case Study: Intralobar Pulmonary Sequestration with Actinomycosis (part 2), colback=white, colframe=orange, width=\textwidth, breakable]
\textbf{Gemini-2.5-Pro:}

\begin{tabular}{@{}lll@{}}
1. \colorbox{green!30}{Intralobar Pulmonary Sequestration} &
2. Pulmonary Actinomycosis \\
3. Lung Abscess  &
4. Congenital Pulmonary Airway Malformation \\
5. Bronchiectasis &
6. Pulmonary Nocardiosis \\
7. Necrotizing Pneumonia &
8. Infected Bronchogenic Cyst \\
9. Tuberculosis &
10. Recurrent Aspiration Pneumonia \\
\end{tabular}

\noindent\rule{\linewidth}{0.4pt}

\textbf{Claude-Sonnet-4:}

\begin{tabular}{@{}lll@{}}
1. Pulmonary echinococcosis (hydatid cyst) &
2. Congenital pulmonary airway malformation &\\
3. Bronchogenic cyst &
4. Lung abscess & \\
5. Necrotizing pneumonia with cavitation &
6. \colorbox{green!30}{Pulmonary sequestration} &\\
7. Bronchiectasis &
8. Aspergilloma &\\
9. Tuberculosis with cavitation &
10. Primary immunodeficiency syndrome &\\
\end{tabular}

\noindent\rule{\linewidth}{0.4pt}

\textbf{o1:}

\begin{tabular}{@{}lll@{}}
1. Congenital Pulmonary Airway Malformation (Type 1) &
2. \colorbox{green!30}{Bronchopulmonary Sequestration} &\\
3. Bronchiectasis &
4. Bronchogenic Cyst &\\
5. Pulmonary Alveolar Proteinosis \\
6. Pulmonary Langerhans Cell Histiocytosis &
7. Cystic Metastatic Lung Disease \\
8. Pulmonary Tuberculosis (Cavitary) \\
9. Pulmonary Echinococcosis (Hydatid Cyst) &
10. Lung Abscess & \\
\end{tabular}

\noindent\rule{\linewidth}{0.4pt}

\textbf{GPT-4.1:}

\begin{tabularx}{\textwidth}{@{}lX@{}}
1. Chronic pulmonary aspergillosis (aspergilloma) & 2. Bronchiectasis \\
3. Recurrent bacterial pneumonia & 4. Lung abscess \\
5. Congenital pulmonary airway malformation \\ 6. \colorbox{green!30}{Pulmonary sequestration} \\
7. Cystic lung disease (non-specific) \\ 8. Pulmonary tuberculosis (chronic cavitary form) \\
9. Non-tuberculous mycobacterial infection \\ 10. Primary immunodeficiency (predisposing to recurrent infections) \\
\end{tabularx}

\noindent\rule{\linewidth}{0.4pt}

\textbf{LLaMA-4-Scout:}

\begin{tabular}{@{}lll@{}}
1. Lymphangioleiomyomatosis & 
2. Pulmonary Lymphangiectasis \\ 
3. Bronchiectasis &
4. Cystic Fibrosis \\
5. Pneumonia & 
6. Interstitial Lung Disease \\
7. Lymphocytic Interstitial Pneumonia & 
8. Pulmonary Alveolar Proteinosis \\
9. Tuberous Sclerosis Complex &
10. Pulmonary Cystic Disease \\
\end{tabular}

\noindent\rule{\linewidth}{0.4pt}

\textbf{Lingshu-32B:}

\begin{tabular}{@{}llll@{}}
1. Pneumonia & 2. Cystic Lung Disease & 3. Aspiration Pneumonia \\ 4. Bronchiectasis &
5. Lung Abscess & 6. Pulmonary Tuberculosis \\ 7. Pulmonary Fungal Infection  &
8. Pulmonary Embolism & 9. Pulmonary Hypertension \\ 10. Pulmonary Fibrosis & \\
\end{tabular}

\noindent\rule{\linewidth}{0.4pt}

\textbf{GLM-4.1V-9B-Thinking:}

\begin{tabular}{@{}lll@{}}
1. Pulmonary Langerhans cell histiocytosis &
2. Cavitary lung abscess \\
3. Bronchogenic cyst &
4. Necrotizing pneumonia \\
5. Pulmonary tuberculosis &
6. Lymphangioleiomyomatosis \\
7. Squamous cell carcinoma with cavitation &
8. Granulomatosis with polyangiitis \\
9. Cystic fibrosis &
10. Invasive aspergillosis \\
\end{tabular}

\noindent\rule{\linewidth}{0.4pt}

\textbf{MedGemma-4B:}

\begin{tabular}{@{}lll@{}}
1. Cystic Fibrosis &
2. Bronchiectasis \\
3. Aspergilloma &
4. Tuberculosis \\
5. Sarcoidosis &
6. Hypersensitivity Pneumonitis \\
7. Bronchiolitis Obliterans &
8. Lymphangioleiomyomatosis (LAM) \\
9. Pulmonary Alveolar Proteinosis (PAP) &
10. Cryptogenic Organizing Pneumonia (COP) \\
\end{tabular}
\end{tcolorbox}
}
    \caption{A case study of MedReaMM. While strong models like Gemini-2.5-Pro correctly identified the primary disease, none recognized the actinomycosis co-infection. This reveals a common limitation: models often prioritize obvious cues while missing critical but subtle clinical findings.}
    \label{fig:case}
\end{figure}

\begin{figure}
    \centering
    \resizebox{0.85\textwidth}{!}{%
    \begin{tcolorbox}[title=Content Extraction, colback=white, colframe=black, width=\textwidth, breakable]
    You are a strict medical information extraction agent. Your task is to extract structured, objective information about ONE real patient described in a case report. The input is the full text of a published case report (converted from PDF).

You must strictly follow these rules:
\begin{enumerate}
    \item[1.] Extract only objective, pre-treatment information relevant to diagnosis, including:
    \begin{itemize}
        \item Basic patient details: age, sex, ethnicity, relevant background
        \item Medical history
        \item Presenting symptoms
        \item Physical examination findings
        \item Laboratory results (pre-treatment only)
        \item Imaging findings (pre-treatment only)
        \item Other diagnostic tests (e.g., endoscopy, biopsy, ECG, etc. — only before treatment)
    \end{itemize}
    Do NOT include any treatment details, post-treatment findings, or clinical outcomes. Only extract what was observed before treatment began.
    \item[2.] Clearly separate laboratory and other diagnostic test results. If any results are presented in a figure or table, you must:
    \begin{itemize}
        \item Summarize the main findings
        \item Include any basic contextual details mentioned in the article for that specific test, such as:Timing of the test (e.g., “on admission”, “hospital day 3”); Test modality (e.g., CT, MRI, pathology, endoscopy); Technical parameters relevant to the modality:For CT or MRI: scan region, contrast usage, slice thickness, etc., For pathology slides: staining method (e.g., H\&E, IHC), magnification level, For endoscopy: scope type, insertion depth, region observed, For ECG: lead configuration, duration of recording, etc.
    \end{itemize}
    If the selected figure includes multiple subfigures, summarize the findings and technical details for each subfigure separately, using labels like (A), (B-C), etc., to match how they are referenced in the paper.
    Summarize this information in natural English sentences, and cite the source in parentheses, such as: “Figure 2”, “Table 1 (Row: Patient A)”.
    \item[3.] Refer to figures/tables like this: Entire figure/table: "Figure 1", "Table 2"; Partial: "Figure 3 (Subfig: A)", "Table 4 (Row: Patient 2)"
    \item[4.] If the case report involves multiple patients, select and extract only ONE.
    \item[5.] At the end, extract the final diagnosis for this patient, exactly as stated in the paper or with minimal paraphrasing.
    \item[6.] Format your output as a JSON object like this:
    \{
  "basics": "Objective patient info, symptoms",
  "supplements": \{
    "Figure 1": "(A) MRI scan performed on the admission day, coronal view, T1-weighted sequence. (B) Chest CT, axial view. (C) Liver biopsy, H\&E staining, 400× magnification, Ki-67=60\%.(D) Laboratory results (E) Clinical photograph.(F) Genetic analysis",
    "Table 2 (Column: Case 1)": "Blood test results show that...",
    "Figure 3 (Subfig: B)": "CT results show..."
  \},
  "Diagnosis": "Final diagnosis stated in the paper"
\}

Important constraints:
\begin{itemize}
    \item All keys in the "supplements" field must begin with "Figure" or "Table"
    \item The "basics" field must be written as a coherent paragraph in natural English, not as a key-value list
\end{itemize}
\end{enumerate}
All outputs must be written in English.

---

Now proceed with the following input: [PDF]

\end{tcolorbox}
}
    \caption{Content extraction prompt.}
    \label{fig:prompt_content_extraction}
\end{figure}

\begin{figure}
    \centering
    \resizebox{0.9\textwidth}{!}{%
    \begin{tcolorbox}[title=Remove Findings From Detailed Caption, colback=white, colframe=black, width=\textwidth, breakable]
    You will be given a detailed medical figure caption (referred to as caption A) that describes one or more subfigures related to medical imaging or testing. These captions may include subjective interpretations, diagnostic conclusions, and descriptions of visual markers (e.g., “arrow indicates…”, “asterisk shows…”).
    
    Your task is to generate a concise, objective caption B that summarizes only the required information from caption A.
    
    For each subfigure (if present), extract and include only the following elements:
    \begin{itemize}
        \item (if available) Subfigure label in parentheses, e.g., (A)
        \item (if available) Time of the examination
        \item Name of the examination, technique, graph or table (e.g., MRI, CT, H\&E staining, laboratory test, clinical photograph, genetic analysis etc.)
        \item (if available) Anatomical region (e.g., brain, liver, lung)
        \item (if available) Viewing plane or orientation (e.g., axial view, sagittal section)
        \item (if available) Staining method (e.g., H\&E, PAS, IHC)
        \item (if available) Magnification level (e.g., 10×, 400×)
        \item (if available) Specific modality/technology (e.g., T2-weighted, DWI, FISH)
        \item (if available) Quantitative values or classifications associated with the examination (e.g., SUVmax=22.4, ADC ratio=0.8, Ki-67=30\%, Grade 2, Stage III)
        
    \end{itemize}
Rules:

    \begin{itemize}
        \item Do NOT include any subjective interpretations or diagnostic statements (e.g., “suggests…”, “indicates abnormality”, "mutation results", “reveals pathology” or "shows" etc.).
        \item Do NOT include any description of visual annotations or labels (e.g., “arrow points to…”, “asterisk indicates…”).
        \item If two or more subfigures correspond to exactly the same examination (same modality, region, orientation, and metadata), you may merge their labels.
        \item Only use information explicitly present in caption A; do not make assumptions or add inferred content.
    \end{itemize}
        
    Output format:
    \begin{itemize}
        \item For multi-panel figures, summarize each subfigure separately using the above structure, and join them in a single paragraph. Example:(A) MRI scan performed on the admission day, coronal view, T1-weighted sequence. (B) Chest CT, axial view. (C) Liver biopsy, H\&E staining, 400× magnification, Ki-67=60\%.(D) Laboratory results (E) Clinical photograph.(F) Genetic analysis
        \item For single-panel figures, write a single concise sentence with no label (e.g., “CT of the abdomen, axial view, performed on day 7.”).
        If the task cannot be completed due to missing, ambiguous, or purely interpretive content, output Error: [reason] explaining why the caption could not be generated.
    \end{itemize}
    
    Now, process this input as caption A:[detailed\_caption]

\end{tcolorbox}
}
    \caption{Remove findings from detailed caption prompt.}
    \label{fig:prompt_remove}
\end{figure}

\begin{figure}
    \centering
    \begin{tcolorbox}[title=Main Evaluation, colback=white, colframe=black, width=\textwidth, breakable]
    \textbf{System Prompt:}
    
You are a medical expert. Given the patient’s demographic information, chief complaint, medical history, and results from various examinations, your task is to identify possible diagnoses. Please enumerate the top 10 most likely diagnoses in order, with the most likely disease listed first. Each item in the list must represent a single, independent disease. If the patient has multiple diseases or complications, please list them separately. Output only under the heading '\#\#\# Output \#\#\#'.

\textbf{User Prompt Content:}

Medical Records ($R$)

Figure n ($I_n$): caption n ($C_n$)

Please output your final answer in the following format:

\#\#\# Output \#\#\#

["Diagnosis A", "Diagnosis B"]
    
\end{tcolorbox}
    \caption{Main evaluation prompt.}
    \label{fig:prompt_main}
\end{figure}

\begin{figure}
    \centering
    \begin{tcolorbox}[title=Image Understanding Test, colback=white, colframe=black, width=\textwidth, breakable]
    \textbf{System Prompt:}

You are a medical assistant.

\textbf{User Prompt Content:}

[Image ($I$)]

For this image, a brief description is as follows: [Caption ($C$)]. 
Please carefully examine the image and provide a detailed description 
of both the visible contents and the information that can be inferred.

\end{tcolorbox}
    \caption{Image understanding test prompt.}
    \label{fig:prompt_image_understanding}
\end{figure}

\begin{figure}
    \centering
    \begin{tcolorbox}[title=Image Understanding Evaluation, colback=white, colframe=black, width=\textwidth, breakable]
    \textbf{System Prompt:} \\
You are a medical image analysis evaluator.

\textbf{User Prompt:}

You are a medical image analysis evaluator. Given a model's description of a medical image, your task is to score the quality of this description based on three criteria:

\begin{enumerate}
  \item[1.] \textbf{Basic Context Recognition}: Does the model correctly identify the type and context of the image? (e.g., whether it is a CT scan of the chest, an ultrasound of the abdomen, or a table of lab results.)
  \item[2.] \textbf{Key Feature Detection}: Does the model correctly describe the key visual features or important data shown in the image?
  \item[3.] \textbf{Clinical Reasoning from Image}: Based on the image and its interpretation, does the model make reasonable clinical inferences?
\end{enumerate}

You are given: \\
\textbf{- Model’s response:} \texttt{\{pred\}} \\
\textbf{- Ground-truth description:} \texttt{\{gt\}}

Please score each criterion from 1 (completely incorrect) to 5 (fully correct). Then provide a brief explanation.

\textbf{Required output format (use this \emph{exact} structure):}
\begin{enumerate}
  \item[1.] Basic Context Recognition: [SCORE]
  \item[2.] Key Feature Detection: [SCORE]
  \item[3.] Clinical Reasoning from Image: [SCORE]
  \item[4.] Explanation: [A brief explanation, 1--3 sentences]
\end{enumerate}
    
\end{tcolorbox}
    \caption{Image understanding evaluation prompt.}
    \label{fig:prompt_image_understanding_eval}
\end{figure}

\begin{figure}
    \centering
    \begin{tcolorbox}[title=Disease Knowledge Test, colback=white, colframe=black, width=\textwidth, breakable]
    \textbf{System Prompt:} \\
You are a medical assistant.

\textbf{User Prompt:}

You are a professional medical assistant. Given the name of a disease, please provide its detailed medical information in the following \textbf{strict format}, using complete sentences.

Disease: \texttt{\{disease\_name\}}

\textbf{Required Output Format (strictly follow this structure):}
\begin{enumerate}
  \item[1.] \textbf{Cause}: Description of the etiology or cause of the disease.
  \item[2.] \textbf{Symptoms}: List of possible symptoms, in natural language.
  \item[3.] \textbf{Diagnostic Tests}: List and describe the key tests needed to diagnose the disease.
  \item[4.] \textbf{Treatment}: Summary of the standard or most effective treatments, including medications, procedures, or lifestyle changes.
\end{enumerate}

Do not include any additional comments, explanation, or formatting outside this structure.
\end{tcolorbox}
    \caption{Disease knowledge test prompt.}
    \label{fig:prompt_disease_kowledge}
\end{figure}

\begin{figure}
    \centering
    \begin{tcolorbox}[title=Disease Knowledge Evaluation, colback=white, colframe=black, width=\textwidth, breakable]
    \textbf{System Prompt:} \\
You are a medical expert.

\textbf{User Prompt:}

You are a medical expert evaluating the quality of a predicted medical summary based on four criteria: \textbf{Cause, Symptoms, Diagnostic Tests, and Treatment}. You will be given both the predicted and ground truth (GT) descriptions for each category. Your task is to compare them and assign a score from 1 to 4 for each category, based on the following rubric:

\textbf{Scoring Rubric (for each field):}
\begin{itemize}
  \item 1 — Incorrect or completely unrelated to GT.
  \item 2 — Partially correct, but missing important elements or contains factual errors.
  \item 3 — Mostly correct, some minor omissions or inaccuracies.
  \item 4 — Fully correct and clinically appropriate.
\end{itemize}

Please carefully assess the semantic and medical accuracy of each predicted field.

\textbf{Disease:} \texttt{\{disease\}}

\begin{itemize}
  \item \textbf{1. Cause} \\
  GT: \texttt{\{gt['Cause']\}} \\
  PRED: \texttt{\{pred['Cause']\}}
  
  \item \textbf{2. Symptoms} \\
  GT: \texttt{\{gt['Symptoms']\}} \\
  PRED: \texttt{\{pred['Symptoms']\}}

  \item \textbf{3. Diagnostic Tests} \\
  GT: \texttt{\{gt['Diagnostic Tests']\}} \\
  PRED: \texttt{\{pred['Diagnostic Tests']\}}

  \item \textbf{4. Treatment} \\
  GT: \texttt{\{gt['Treatment']\}} \\
  PRED: \texttt{\{pred['Treatment']\}}
\end{itemize}

\textbf{Output Format (only output this JSON dictionary):}

\{

  "Cause": score from 1 to 4,
  
  "Symptoms": score from 1 to 4,
  
  "Diagnostic Tests": score from 1 to 4,
  
  "Treatment": score from 1 to 4
  
\}

Do not explain your answer.

\end{tcolorbox}
    \caption{Disease knowledge evaluation prompt.}
    \label{fig:prompt_disease_eval}
\end{figure}

\begin{figure}
    \centering
    \begin{tcolorbox}[title=Ground Truth Reasoning Path Extraction, colback=white, colframe=black, width=\textwidth, breakable]
\textbf{System Prompt:} \\
You are an expert medical diagnostician.

\textbf{User Prompt Content:} \\
Please provide a detailed step-by-step reasoning path for the following medical case. Given the patient’s demographic information, chief complaint, medical history, and results from various examinations, your task is to reason through the case and arrive at possible diagnoses. Clearly outline your thought process, including relevant findings, differential diagnoses considered, and the rationale for prioritizing certain conditions.

Output only under the heading \texttt{\#\#\# reasoning path \#\#\#}.

\textbf{INPUT:} [PDF]

\textbf{Expected Output Format:}

\#\#\# reasoning path \#\#\#

[Step 1: ..., Step 2: ..., ..., Conclusion: ...]

\end{tcolorbox}
    \caption{Ground truth reasoning path extraction prompt.}
    \label{fig:prompt_reasoning_path_extraction_gt}
\end{figure}

\begin{figure}
    \centering
    \begin{tcolorbox}[title=Reasoning Path Extraction, colback=white, colframe=black, width=\textwidth, breakable]
\textbf{System Prompt:} \\
You are an expert medical diagnostician. Please provide a detailed step-by-step reasoning path for the following medical case.

\textbf{Patient Information:} 

Medical Records (R)

\textbf{Available Medical Data:}

Figure n ($I_n$): caption n ($C_n$)

\textbf{Reference Diagnosis:} 

Diagnosis ($D$)

\textbf{Please be thorough and demonstrate expert-level medical reasoning.}

\textbf{Output Format:} \\
\#\#\# reasoning path \#\#\#

[Step 1: ..., Step 2: ..., ..., Conclusion: ...]

\end{tcolorbox}
    \caption{Reasoning path extraction prompt.}
    \label{fig:prompt_reasoning_path_extraction}
\end{figure}

\begin{figure}
    \centering
    \begin{tcolorbox}[title=Reasoning Path Evaluation, colback=white, colframe=black, width=\textwidth, breakable]
    \textbf{System Prompt:} \\
You are a medical expert.

\textbf{User Prompt Content:} \\
Please, based on the reasoning path extracted from the article, evaluate the reasoning path generated by the model. Please use the following medical standards.

\textbf{Evaluation Criteria:}

\begin{enumerate}
    \item \textbf{Reasoning Quality (Primary):} Assess logical flow, medical accuracy, depth of analysis, completeness of differential diagnosis, evidence integration, and clinical sophistication. Demand the highest possible standards.
    \item \textbf{Diagnosis Accuracy (Secondary):} Determine if the ground truth diagnosis is present within the top 5 generated diagnoses.
\end{enumerate}

\textbf{Scoring Rubric (1–5):}

\begin{itemize}
    \item \textbf{5 (Excellent):} Perfect logical flow, comprehensive differential diagnosis with detailed pathophysiological reasoning, expert integration of all clinical data, sophisticated medical insights.
    \item \textbf{4 (Good):} Medically sound and well-structured reasoning with good depth, but has noticeable limitations: may miss some important differentials, lack some pathophysiological depth, or contain areas of superficial analysis. Solid but not exceptional.
    \item \textbf{3 (Adequate):} Shows basic medical competence but has clear limitations: missing key differentials, superficial analysis, incomplete data integration, gaps in logical flow, or lack of sophisticated insights. Represents standard competency level.
    \item \textbf{2 (Below Standard):} Significant deficiencies: major gaps in differential diagnosis, poor logical flow, superficial analysis, medical inaccuracies, or incomplete reasoning.
    \item \textbf{1 (Inadequate):} Fundamentally flawed, medically incorrect, or completely inadequate.
\end{itemize}

Reference reasoning path: [Ref.]

Model predicted reasoning path: [Pred.]

\end{tcolorbox}
    \caption{Reasoning path evaluation prompt.}
    \label{fig:prompt_reasoning_path_eval}
\end{figure}

\begin{figure}
    \centering
    \begin{tcolorbox}[title=Prompt Robustness (prompt 2), colback=white, colframe=black, width=\textwidth, breakable]
    \textbf{System Prompt:}

You are an experienced medical doctor with expertise in differential diagnosis.

Carefully analyze the provided patient information including demographics, chief complaint, medical history, and examination results.

Consider all possible conditions and use your medical knowledge to systematically evaluate each potential diagnosis.

Provide the top 10 most likely diagnoses ranked by probability, with the most probable condition first.

Each diagnosis should be a distinct medical condition. If multiple conditions are present, list them separately.

Format your response under the heading \texttt{\#\#\# Output \#\#\#}.

\textbf{User Prompt Content:}

Medical Records ($R$)

Figure n ($I_n$): caption n ($C_n$)

Please output your final answer in the following format:

\#\#\# Output \#\#\#

["Diagnosis A", "Diagnosis B"]

\end{tcolorbox}
    \caption{Prompt 2 used in prompt robustness analysis.}
    \label{fig:prompt_2}
\end{figure}

\begin{figure}
    \centering
    \begin{tcolorbox}[title=Prompt Robustness (prompt 3), colback=white, colframe=black, width=\textwidth, breakable]
    \textbf{System Prompt:}

You are a medical expert practicing evidence-based medicine.

Analyze the patient case using clinical evidence and established diagnostic criteria.

Evaluate each potential diagnosis based on the strength of supporting clinical evidence.

Consider prevalence, clinical presentation patterns, and diagnostic probability.

List the top 10 most likely diagnoses ordered by clinical evidence strength and probability.

Each diagnosis should be a distinct medical entity.

Format your response under \texttt{\#\#\# Output \#\#\#}.

    \textbf{User Prompt Content:}

Medical Records ($R$)

Figure n ($I_n$): caption n ($C_n$)

Please output your final answer in the following format:

\#\#\# Output \#\#\#

["Diagnosis A", "Diagnosis B"]
\end{tcolorbox}
    \caption{Prompt 3 used in prompt robustness analysis.}
    \label{fig:prompt_3}
\end{figure}

\twocolumn
\label{sec:appendix}

\end{document}